\newif\ifneurips
\neuripstrue   

\ifneurips
  \documentclass{article}
    \usepackage[preprint]{neurips_2025}
\else
  \documentclass[11pt]{article}
  \usepackage[margin=1.25in]{geometry}
\fi

\usepackage{lmodern}

\usepackage{amsmath,amssymb,amsthm}
\usepackage{booktabs}
\usepackage{hyperref}
\usepackage{xcolor}
\usepackage{tcolorbox}
\usepackage{comment}
\usepackage{titletoc}
\usepackage{tcolorbox}

\usepackage[table]{xcolor}
\definecolor{FlexokiBlue}{HTML}{C6DDE8}

\usepackage{adjustbox}
\usepackage{booktabs,tabularx}
\tcbset{
  colback=blue!8!white,
  colframe=blue!8!white, 
  boxrule=0pt,
  arc=4pt,
}
\definecolor{warmivory}{HTML}{FFFBED}


\newtheorem{proposition}{Proposition}
\newtheorem*{proposition*}{Proposition}
\newtheorem{lemma}{Lemma}
\newtheorem*{definition*}{Definition}

\newtheorem*{corollary*}{Corollary}
\theoremstyle{remark}

\newtheorem*{remark*}{Remark}
\definecolor{flexokilavander}{HTML}{A699D0}
\colorlet{flexokilavanderlight}{flexokilavander!18!white}

\definecolor{figureblue}{HTML}{203B73}

\newcommand{\ind}[1]{\mathbf{1}_{\{#1\}}}
\newcommand{\Nt}{N_t}
\newcommand{\dH}{d_H}
\newcommand{\Ndata}{\mathcal{N}(D)}
\newcommand{\Ntk}{N_{t,k}}

\title{A Boltzmann Kernel for Discrete Diffusion: How Locality and Equivariance Explain Creativity}

\author{%
  \parbox{\textwidth}{\centering
Alberto Bechara\textsuperscript{1,2} \thanks{Corresponding author: abecharabordes@gmail.com},
    Navonil Neogi\textsuperscript{3},
    Nabil Iqbal\textsuperscript{1},\\[0.5em]
    Erik J. Bekkers\textsuperscript{1},
    Maria Esteban-Casadevall\textsuperscript{1}\\[0.75em]
\adjustbox{max width=\linewidth}{%
  \normalfont
  \textsuperscript{1}University of Amsterdam\quad
  \textsuperscript{2}CFIS, Universitat Politècnica de Catalunya\quad
  \textsuperscript{3}Durham University%
}
  }%
}

\begin{document}

\maketitle
\begin{abstract}
Despite the strong generative capabilities of discrete diffusion models, the mechanisms that enable them to generalize beyond their training data remain poorly understood. We study this question through the architectural inductive biases of locality and translation equivariance. For Uniform Discrete Diffusion Models (UDMs), we derive an analytical characterization of the optimal denoiser under these constraints. The resulting predictor takes the form of a Boltzmann-like kernel over training patches, weighting each patch exponentially by its Hamming distance to the noisy input. Experiments on MNIST and CIFAR-10 show close agreement between trained models and the analytical prediction, with discrepancies increasing as receptive-field and vocabulary sizes reduce local data coverage. We further extend our analysis to Masked Diffusion Models (MDMs), where we introduce a relaxed kernel formulation that captures their empirical generalization behaviour, and validate it experimentally. 
\end{abstract}

\section{Introduction}
\label{sec:intro}

Even with the remarkable progress of generative models, the mechanisms underlying their creative capabilities remain only partially understood. In the continuous-data setting, considerably more is known: generalizable features have been linked to model-class inductive biases, signals arising from components of the data distribution, and architectural choices~\cite{kamb,lukoianov, kadkhodaie}. In the discrete setting, however, our understanding remains  limited. In this work, we focus on discrete diffusion models and ask the following question:  
\begin{tcolorbox}[
    colback=figureblue!10!white,
    colframe=figureblue,
    boxrule=0pt,
    leftrule=1.5pt,
    arc=0pt,
    outer arc=0pt,
    left=10pt,
    right=10pt,
    top=6pt,
    bottom=6pt
]
\centering
\emph{How can the creative capabilities of discrete diffusion models
be explained by their architectural inductive biases?}
\end{tcolorbox}


Following the framework of Kamb \& Ganguli \cite{kamb}, we treat \emph{inductive bias} as a primary source of generalization, and hence of creative behaviour. In the absence of such constraints, attaining the global optimum of the training objective would, in principle, allow the model to reproduce the empirical data distribution and therefore memorize the training set. This is not, however, the behaviour typically observed in trained models. Architectural inductive biases restrict the model distribution $p_\theta$ to a smaller function class $\mathcal{F}$, making the unconstrained global optimum generally unattainable. Training instead approaches the optimum within $\mathcal{F}$, which can deviate systematically from the empirical data distribution and thereby induce generalization beyond the training samples. Characterizing this constrained optimum analytically therefore provides a way to understand the creative behaviour induced by the underlying architecture.

We study two fundamental architectural inductive biases. The first is \emph{locality}, where the model is given a restricted receptive field and can condition only on a subset of the sample tokens at a time. The second is \emph{translation equivariance}, which constrains translated inputs to induce correspondingly translated outputs, as naturally implemented, for example, through weight sharing in convolutional architectures~\cite{cohen2016groupequivariantconvolutionalnetworks}. This leads to the following contributions:

\begin{itemize}
\item We derive the first analytical theory of creativity for uniform discrete diffusion models when parametrized by local and translation-equivariant architectures (Section~\ref{sec:usdm}). Our results extend naturally to other corruption processes, such as the GIDD~\cite{vonrütte2025generalizedinterpolatingdiscretediffusion} framework (Appendix~\ref{app:GIDD}).

\item We experimentally show that the theoretical results hold in practice across different datasets~\cite{mnist,fashionmnist,Krizhevsky} and architectures (CNNs, NATTEN models~\cite{hassani2023neighborhoodattentiontransformer}, and GNNs~\cite{gilmer2017neuralmessagepassingquantum}) (Section~\ref{sec:usdm-experiments}).

\item We show that, for Masked Diffusion, the reverse dynamics alone are insufficient to characterize the model's generalization behavior, even under locality and translation equivariance. We therefore propose a heuristic theory incorporating additional architectural and optimization constraints, and support it with experimental evidence (Section~\ref{sec:mdm}).
\end{itemize}


\section{An analytical expression for Uniform Discrete Diffusion}
\label{sec:usdm}

We work within the framework established by Kamb \& Ganguli~\cite{kamb} and derive an analogous theory for discrete domains. In uniform discrete diffusion models~\cite{hoogeboom2021argmaxflowsmultinomialdiffusion,austin2023structureddenoisingdiffusionmodels,schiff2025simpleguidancemechanismsdiscrete,sahoo2025diffusionduality} a token with state $x_0^{(k)} \in \mathcal{V}$ and index $k$ is noised independently and uniformly: at time $t$ it has probability $\alpha_t$ to stay in its original state  and $1-\alpha_t$ to change to a random state among $K=|\mathcal{V}|$ possible states. 

We train the model as a denoiser that predicts a clean token from a noisy sample by minimizing a loss function $\mathcal{L}(\theta)$. At a global minimum of this objective, the theoretically optimal denoiser recovers the true reverse posterior $q(x_0^{(k)}|x_t)$ (refer to Appendix~\ref{app:memorization} for a proof). In practice, however, the data distribution is only observed through a finite training set $D$ and is therefore approximated by the empirical distribution~\cite{kamb} \[ p_{\mathrm{data}}(x) \approx p_{\mathrm{empirical} }(x)= \frac{1}{|D|} \sum_{x_0 \in D} \delta_{x_0}(x). \] Consequently, the posterior learned at the empirical optimum is only the \textit{empirical posterior} $q_{\mathrm{emp}}(x_0^{(k)}|x_t) = \sum_{x_0'\in D}q_{\mathrm{emp}}(x_0'|x_t)\,\ind{x_0'^{(k)}=x_0^{(k)}}$ with $q_{\mathrm{emp}}(x_0'|x_t)\propto q(x_t|x_0')$, which is fundamentally constrained by the finite support of the empirical data distribution. 

Under this empirical posterior, both uniform and masked diffusion memorize the training set, analogously to the continuous case studied by Kamb \& Ganguli~\cite{kamb}. For uniform diffusion in particular we find the empirical posterior to be a Boltzmann distribution over the training samples (Appendix \ref{app:memorization})
\[
q_{\mathrm{emp}}(x_0'\mid x_t)=\frac{e^{-\beta_t\,\dH(x_t,x_0')}}{\sum_{x_0''\in D}e^{-\beta_t\,\dH(x_t,x_0'')}},
\qquad
\beta_t=\log\Big(1+\frac{\alpha_t K}{1-\alpha_t}\Big),
\]
where $\dH$ is the Hamming distance, $d_H(x, y) := \sum_{i=1}^{d} \ind{x^{(i)} \neq y^{(i)}}$. As $t\to0$, $\beta_t\to\infty$ and the posterior concentrates on the training sample nearest to $x_t$. Now assuming we use ancestral sampling \cite{austin2023structureddenoisingdiffusionmodels} we find that the probability of not memorising is given by the following bound, which tends to 0 with bigger time discretisations:

\begin{tcolorbox}[
    colback=figureblue!10!white,
    colframe=figureblue,
    boxrule=0pt,
    leftrule=1.5pt,
    arc=0pt,
    outer arc=0pt,
    left=10pt,
    right=10pt,
    top=6pt,
    bottom=6pt
]
\begin{proposition}[Uniform diffusion memorizes]
\label{prop:udm-memorization-main}
Let $x_0$ be generated by ancestral sampling with the empirical optimum on a time grid $1=t_N>\dots>t_1>t_0=0$, and let $\delta_D$ be the minimum Hamming distance between two training samples. Then
\[
\mathbb P\big(x_0\notin D\big)
\;\le\;
\frac{d^2}{2}\sum_{n=2}^{N} h_n^2
\;+\;
\frac{4d\,\rho}{\delta_D}
\;+\;
d\,(|D|-1)\,e^{-\beta_{t_1}\delta_D/2},
\]
where $\rho=(1-\alpha_{t_1})\frac{K-1}{K}$, $h_n=\frac{\alpha_{t_{n-1}}-\alpha_{t_n}}{\alpha_{t_{n-1}}(1-\alpha_{t_n})}$ for $n<N$, and $h_N=\alpha_{t_{N-1}}$. The right-hand side can be made arbitrarily small by refining the time grid.
\end{proposition}
\end{tcolorbox}

The proof is given in Appendix~\ref{App: A.4}. We also run experiments that empirically verify memorization to be the case in practice: where every run finishes with more than $98.8\%$ of its tokens equal to those of the closest training sample for low vocabulary experiments $K \in \{2,4,8\}$. 

Given that the empirical posterior overfits, we hypothesize following Kamb $\&$ Ganguli \cite{kamb} that there must be some kind of inductive bias stopping the model from converging to this optimal of the loss function after training. We now investigate how architectural inductive biases, in particular locality and equivariance, constrain the denoiser and drive it away from the memorizing empirical posterior.

\textbf{Equivariance and locality break memorization.} 

Locality restricts the information available to the denoiser at each token position. Rather than conditioning on the entire noisy sample $x_t$, the prediction at position $k$ depends only on a local neighborhood
$\Ntk \subseteq x_t$ centered at $k$, whose shape and size may be arbitrary, but for simplicity we will consider from now on only square patches of size $w\times w$. This restriction leads to the following proposition:

\begin{figure}[ht!]
\centering
\includegraphics[width=1.\textwidth]{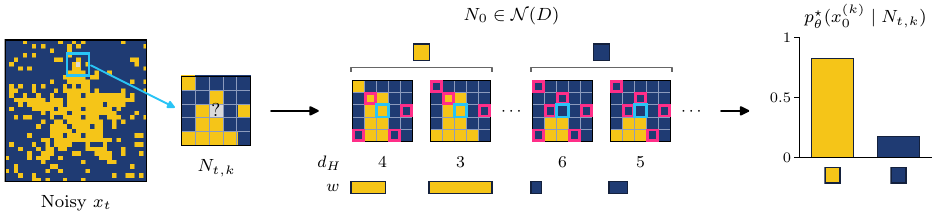}
\caption{The Boltzmann weights depend on the Hamming distance between the noisy patch, coming from the noisy image $x_t$, and the patches $N_0$ coming from the training samples $D$. The Hamming distance measures the number of tokens that disagree with the noisy patch (highlighted in pink). }
\label{fig:1}
\end{figure}



\begin{tcolorbox}[
    colback=figureblue!10!white,
    colframe=figureblue,
    boxrule=0pt,
    leftrule=1.5pt,
    arc=0pt,
    outer arc=0pt,
    left=10pt,
    right=10pt,
    top=6pt,
    bottom=6pt
]
\begin{proposition}[Locality]
\label{prop:locality}
If a Uniform Diffusion model $p_\theta$ is constrained to observe only the local patch $N_{t,k}$, the training objective forces the optimal local predictor $p_\theta^*$ to approximate the local marginal posterior:
\begin{equation}
\label{eq:udm-local}
p_\theta^*\left(x_0^{(k)} \mid x_{t}\right)= q\left(x_0^{(k)} \mid N_{t,k}\right)
\end{equation}
where $q(x_0^{(k)} \mid N_{t,k})$ is the true reverse token distribution conditioned on a patch $N_{t,k}$ centered at token $k$. 
\end{proposition}
\end{tcolorbox}
The second inductive bias that we consider is \textit{group equivariance}. This means we architecturally restrict the class of models further to those which follow, for some group $G$:

\begin{equation}
p_\theta\big(x_0^{(k)} \mid x_t\big) = p_\theta\big(x_0^{(g\cdot k)} \mid g\cdot x_t\big), \qquad \forall\, g \in G,
\label{eq:equivariance}
\end{equation}

Our main result holds for any surjective group action (see Appendix~\ref{app:theory-udm}). For simplicity, we focus on the translation group action.

\begin{tcolorbox}[
    colback=figureblue!10!white,
    colframe=figureblue,
    boxrule=0pt,
    leftrule=1.5pt,
    arc=0pt,
    outer arc=0pt,
    left=10pt,
    right=10pt,
    top=6pt,
    bottom=6pt
]
\begin{proposition}[Translation equivariance] \label{prop:main}
For a local-equivariant Uniform Diffusion model following Eq.~\eqref{eq:udm-local} and Eq.~\eqref{eq:equivariance}, the loss optimum is given by the unanchored local posterior:

\begin{equation}
p_\theta^{\star}\big(x_0^{(k)} \mid x_t\big) = q_{}\big(x_0^{(k)} \mid \Nt\big).
\label{eq:local-equiv-general}
\end{equation}
The local equivariant UDM analytical loss optimum then has the following form:
\begin{equation}
\label{eq:optimum}
\small
p_\theta^{\star}\big(x_0^{(k)}=c \mid \Nt\big)
=
\frac{
    \sum_{N_0 \in \Ndata}
    \ind{\mathrm{center}(N_0){=}c}\,
    e^{-\beta_t \dH(N_0,\Nt)}
}{
    \sum_{N_0' \in \Ndata}
    e^{-\beta_t \dH(N_0',\Nt)}
},
\quad
\beta_t =
\log\!\left(1+\frac{\alpha_t K}{1-\alpha_t}\right)
\end{equation}
\end{proposition}
\end{tcolorbox}

where $\Nt$ now denotes the coordinate-free window shapes. The set $\Ndata$ contains all $w\times w$ patches extracted from every spatial location of every training sample, with repeated patches appearing according to their empirical frequency, and $\dH$ denotes the Hamming distance between two patches of side $w$.

\textbf{Interpretation of analytical expression.} Equation~\ref{eq:optimum} provides a precise characterization of the generalization mechanism of the optimal denoiser beyond the training samples. Predictions are determined by comparisons between the noisy input patch and patches observed in the training set. As denoising approaches its final stages, this weighting becomes increasingly selective: since $\beta_t \to \infty$ as $\alpha_t \to 1$ (equivalently, $t \to 0$), the prediction becomes dominated by training patches that are closest to the current neighborhood. Sample-level memorization is turned into similarity-weighted patch matching, effectively reducing the combinatorial space of samples from $K^{H\cdot W}$ to $K^{w^2}$ (locality) and amplifying the dataset signal to delocalized samples (equivariance). This  analytical explanation of creativity will decay in small datasets over large $K$ or large $w$, effectively returning to the global posterior at $w=\min(H, W)$. Figure~\ref{fig:1} illustrates how the Boltzmann weights from Proposition~\ref{prop:main} are defined.



\section{Experiments}
\label{sec:usdm-experiments}

Equation~\ref{eq:optimum} gives the minimum achievable loss under these constraints. We claim that trained local-equivariant UDMs will converge to this optimum. We test this hypothesis in controlled experiments. 

To do that, we compare the trained CNN with the analytical prediction in Eq.~\eqref{eq:optimum} on MNIST and CIFAR-10, with vocabulary sizes $K\in\{2,4,8\}$. The results are reported in Table~\ref{Table 1}. For small windows $w$, the empirical results closely match the analytical expression across all datasets. As the window size increases, however, the discrepancy between the analytical prediction and the experimental results becomes larger, with the gap growing for larger values of $K$. This behavior is consistent with the $K^{w^2}$ coverage scaling analyzed in Section~\ref{sec:usdm}. In Appendix~\ref{app:exp}, we further show that this gap decreases as the dataset size increases, along with definitions of the evaluation metric, details of the experimental setup and ablations on other architectures.


\begin{figure}[h]
\centering
\includegraphics[width=1.\textwidth]{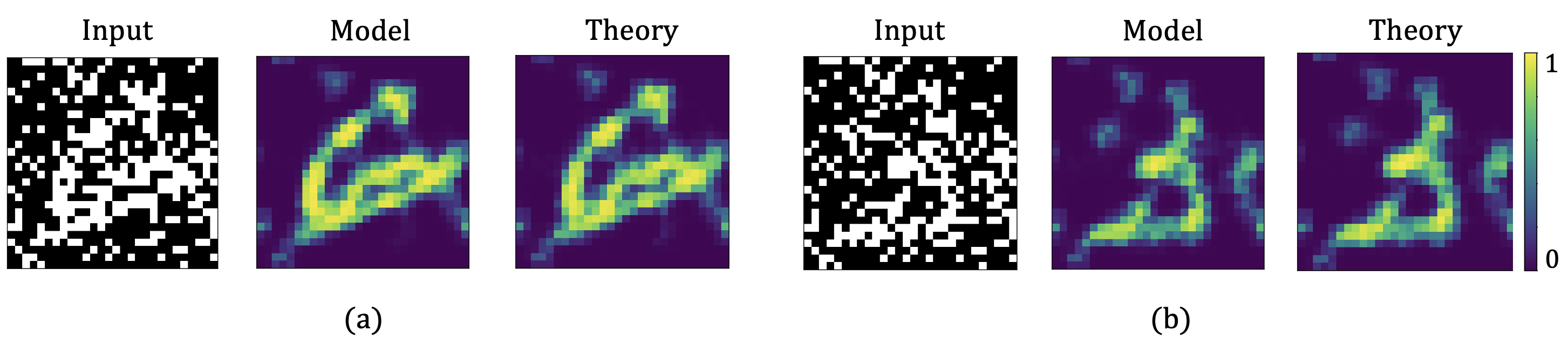}
\caption{Model predictions vs analytical prediction from Eq.~\eqref{eq:optimum}, on the MNIST dataset with $K=2$, $w{=}5$ and $\alpha_t{=}0.5$. }
\label{fig:usdm-heatmap}
\end{figure}

\begin{table}[ht!]
\centering
\renewcommand{\arraystretch}{1.15}
\resizebox{\textwidth}{!}{%
\begin{tabular}{lccccccccc}
\toprule
& \multicolumn{3}{c}{
    $w{=}3$}
& \multicolumn{3}{c}{
   $w{=}5$}
& \multicolumn{3}{c}{
    $w{=}7$} \\
\cmidrule(lr){2-4}
\cmidrule(lr){5-7}
\cmidrule(lr){8-10}

\textbf{Dataset}
& KL $\downarrow$ & TV $\downarrow$ & Top-1 $\uparrow$
& KL $\downarrow$ & TV $\downarrow$ & Top-1 $\uparrow$
& KL $\downarrow$ & TV $\downarrow$ & Top-1 $\uparrow$ \\
\midrule

MNIST ($K{=}2$)
& 0.0009 & 0.0065 & 99.56\%
& 0.0057 & 0.0137 & 99.12\%
& 0.0104 & 0.0172 & 98.68\% \\

CIFAR-10 ($K{=}2$)
& 0.0013 & 0.0113 & 98.40\%
& 0.0038 & 0.0170 & 97.52\%
& 0.0144 & 0.0241 & 97.48\% \\

CIFAR-10 ($K{=}4$)
& 0.0060 & 0.0248 & 97.20\%
& 0.0269 & 0.0452 & 95.04\%
& 0.0991 & 0.0740 & 91.80\% \\

CIFAR-10 ($K{=}8$)
& 0.0244 & 0.0514 & 92.24\%
& 0.1382 & 0.1014 & 89.12\%
& 1.6989 & 0.1758 & 82.48\% \\

\bottomrule
\end{tabular}%
}

\vspace{8pt}
\caption{Mean KL, TV distance, and top-1 agreement
between the trained model (CNN) and the analytical expression
in Eq.~\eqref{eq:optimum} for $N{=}5000$ training samples.
Arrows indicate whether lower ($\downarrow$) or higher
($\uparrow$) values are better.}
\label{Table 1}
\end{table}


\newpage
\section{Extension to Masked Diffusion Models}
\label{sec:mdm}

In contrast, in Masked Diffusion~\cite{austin2023structureddenoisingdiffusionmodels,sahoo2024simpleeffectivemaskeddiffusion,shi2025simplifiedgeneralizedmaskeddiffusion} a corrupted token retains no information about its clean value, while every visible token is exactly clean. The resulting optimum (derived in Appendix~\ref{app:GIDD}), which we call the \emph{exact lookup}, can be interpreted as the limit $\beta_t \to \infty$ of Eq.~\eqref{eq:optimum}, with the Hamming distance restricted to the visible positions: the exponential weighting reduces to an indicator of exact agreement on those positions. This gives

\begin{equation}
p_\theta^{\star}\big(x_0^{(k)} \mid x_t\big) = q_{\mathrm{emp}}\big(x_0^{(k)} \mid \Nt^{\mathrm{vis}}\big)
= \frac{\displaystyle\sum_{N_0 \in \Ndata} \ind{N_0^{\mathrm{vis}}{=}\Nt^{\mathrm{vis}}}\,\ind{\mathrm{center}(N_0){=}x_0^{(k)}}}
       {\displaystyle\sum_{N_0' \in \Ndata} \ind{N_0^{\prime\,\mathrm{vis}}{=}\Nt^{\mathrm{vis}}}},
\label{eq:mdm-local-equiv-opt}
\end{equation}

where $\Nt^{\mathrm{vis}}$ denotes the restriction of $\Nt$ to its unmasked (visible) positions. The expression is undefined whenever no training patch $N_0 \in \Ndata$ agrees with $\Nt^{\mathrm{vis}}$ at every visible position.

\begin{table*}[ht!]
\noindent
\begin{minipage}[t]{0.60\textwidth}
\vspace{0pt}

This behaviour does not hold in practice as MDMs are models able to interpolate and generalize \cite{shi2025simplifiedgeneralizedmaskeddiffusion}.
We attribute this discrepancy to the parameterization of the neural network.
In practice we train $p_\theta = \mathrm{softmax}(f_\theta)$, so matching the empirical posterior requires
\[
\mathrm{softmax}(f_\theta)_{x_0^{(k)}} = \sum_{N_0} w(N_0 \mid N_{t,k})\, \mathbf{1}\{N_0^{(k)} = x_0^{(k)}\},
\]
with $w \propto e^{-\beta d_H}$ at $\beta \to \infty$.
The logits must then act as an indicator, $f_\theta \in \{\infty, -\infty\}$.
Finite logits $f_\theta$ are unable to reproduce this extreme behaviour, yielding a continuous relaxation of the exact lookup.
We model this relaxation heuristically using the same exponential form as in Eq.~\ref{eq:optimum}, replacing $\beta_t$ with a fitted parameter $\hat{\beta}$. Unlike in UDM, where $\beta_t$ is determined by the diffusion dynamics, we hypothesize that in MDM $\hat{\beta}$ is determined by the architecture and training process. We discuss this approximation further in Appendix~\ref{app:exp-mdm}.


We test our hypothesis experimentally on MNIST and CIFAR-10 ($K= 2,4,8$). The results are reported in Table~\ref{tab:mdm}. The relaxed heuristic kernel beats the exact lookup (KL exact) at every $w$. As window size increases the exact posterior deteriorates, and for the same reasons as in Section~\ref{sec:usdm-experiments} the match against Eq.~\ref{eq:optimum} decays as well. 
In Appendix~\ref{app:mdm-check} we verify that this agreement is not universal. 

\end{minipage}
\hfill
\begin{minipage}[t]{0.35\textwidth}
\vspace{1.mm}
\centering

\footnotesize
\setlength{\tabcolsep}{2pt}
\renewcommand{\arraystretch}{1.12}

\begin{tabular*}{\linewidth}{
    @{\extracolsep{\fill}}
    l
    c
    c
    c
    c
    @{}
}
\toprule
\textbf{Dataset}
& $w$
& \shortstack{KL\\exact}
& $\hat{\beta}$
& \shortstack{KL at\\$\hat{\beta}$} \\
\midrule
  MNIST & 3 & 0.0011 & 9.6 & 0.0015 \\
      & 5 & 0.0440 & 6.6 & 0.0102 \\
      & 7 & 0.3410 & 2.5 & 0.0141 \\

\addlinespace[2pt]

\shortstack[l]{CIFAR-10\\($K{=}4$)}
& 3 & 0.0301 & 5.2 & 0.0139 \\
& 5 & 1.4868 & 2.8 & 0.0496 \\
& 7 & 2.1015 & 1.8 & 0.0867 \\

\addlinespace[2pt]

\shortstack[l]{CIFAR-10\\($K{=}8$)}
& 3 & 0.7101 & 4.4 & 0.0405 \\
& 5 & 4.2007 & 2.5 & 0.1056 \\
& 7 & 3.3673 & 1.6 & 0.1861 \\

\bottomrule
\end{tabular*}

\vspace{8pt}
\caption{MDM with CNN backbone and $N{=}5000$. KL from the exact lookup (Eq.~\eqref{eq:mdm-local-equiv-opt}) versus relaxed heuristic
(Eq.~\eqref{eq:optimum}), averaged over noise levels and restricted to
queries whose visible context is matched exactly by a training patch.} 
\label{tab:mdm}

\end{minipage}
\end{table*}

\section{Conclusion}
We showed that the creativity exhibited by Uniform Discrete Diffusion models can be fundamentally explained by the analytical mechanism in Eq.~\ref{eq:optimum}, and we introduced heuristics for extending this theoretical perspective to masked diffusion models. As future work, one could extend the analysis to account for correlations across coordinates or for structure within the vocabulary itself. Our analysis is restricted to local, translation-equivariant denoisers. Extending the Boltzmann kernel to capture longer-range data correlations and more general architectures, such as attention-based models, is a natural next step towards a theory of creativity in discrete diffusion. More broadly, our results suggest that in discrete diffusion, as in the continuous case, creativity is not a failure to learn the data but a natural consequence of dynamics and architecture design. 

\bibliographystyle{plain}
\bibliography{references}

\appendix

\newpage
\section*{Appendix}
\startcontents[appendices]
\printcontents[appendices]{}{0}{\setcounter{tocdepth}{2}}
\vspace{1em}

\newpage

\section{The model global optimum memorizes}
\label{app:memorization}

In this section we prove that the empirical global optima in uniform and masked diffusion memorize the dataset. We start by proving that the optimum of the cross-entropy loss recovers the true posterior (Section~\ref{App: A.1}), and then derive the empirical posterior for both masked and uniform corruption (Section~\ref{App: A.2}). We then prove how each posterior leads to memorization at sampling time, first for masked diffusion (Section~\ref{App: A.3}) and then for uniform diffusion (Section~\ref{App: A.4}). Section~\ref{App: A.5} checks the uniform case numerically.

\subsection{The optimum of the cross-entropy loss is the true posterior}  \label{App: A.1}

Both uniform and masked diffusion models are trained with a cross-entropy objective. We therefore begin by showing that the global minimizer of this loss is the true posterior distribution. This is a standard result but we include it for completeness.

\begin{proposition}[The CE loss optimum is the true posterior]
For any forward kernel that corrupts tokens independently, the minimizer
of
\begin{equation}
\mathcal{L}(\theta) = \mathbb{E}_{t\sim U(0,1)}\big[\mathcal{L}_t(\theta)\big],
\qquad
\mathcal{L}_t(\theta) = \mathbb{E}_{q(x_0)}\left[\mathbb{E}_{q(x_t\mid x_0)}\left[\sum_{k\in S_t} -\,\omega(t)\log p_\theta\big(x_0^{(k)}\mid x_t\big)\right]\right]
\label{eq:ce-objective}
\end{equation}
where $S_t\subset x_t$ is a subset of tokens determined by the corruption process, is the true posterior,
\begin{equation*}
p_\theta^{\star}\big(x_0^{(k)} \mid x_t\big) = q\big(x_0^{(k)} \mid x_t\big),
\end{equation*}
at every $t$ and every scored position $k \in S_t$.
\end{proposition}

\begin{proof}
Fix $t$ in Eq.~\eqref{eq:ce-objective}. Since $\omega(t)\ge 0$, minimizing
$\mathcal{L}$ over unrestricted $p_\theta$ amounts to minimizing
$\mathcal{L}_t$ separately for each $t$, since we are using $p_\theta$ independent of $t$. Applying the tower property to $q(x_0)q(x_t\mid x_0)=q(x_t)q(x_0\mid x_t)$ gives
\begin{equation*}
\mathcal{L}_t(\theta)
= \omega(t)\sum_{k=1}^{d}\ \mathbb{E}_{q(x_t)}\!\left[\ind{k\in S_t}\
\mathbb{E}_{q(x_0^{(k)}\mid x_t)}\!\big[-\log p_\theta\big(x_0^{(k)}\mid x_t\big)\big]\right],
\end{equation*}
the inner expectation being over the marginal $q(x_0^{(k)}\mid x_t)$
as the summand depends on $x_0$ only through its $k$-th coordinate.
Because we have a cross-entropy, by Gibbs' inequality
\begin{equation*}
CE\big(q(\cdot\mid x_t),\,p_\theta(\cdot\mid x_t)\big)
= H\big(q(\cdot\mid x_t)\big) + D_{KL} \big(q(\cdot\mid x_t)\,\|\, p_\theta(\cdot\mid x_t) \big)
\end{equation*}
is minimized iff $p_\theta(\cdot\mid x_t)=q(\cdot\mid x_t)$, concluding our proof.
\end{proof}

In the case of UDM, $S_t = x_t$, as every token has plausible corruption and contributes to the loss.

In the case of MDM, $S_t = M_t$, the set of masked tokens at time $t$. Tokens that are already denoised cannot be changed and are assumed to be correct.

\subsection{The empirical posteriors for UDM and MDM} \label{App: A.2}

The posterior $q(x_0\mid x_t)\propto p_{\mathrm{data}}(x_0)\,q(x_t\mid x_0)$ is not tractable since $p_{\mathrm{data}}$ is unknown. However, given a finite training set $D$ one can approximate the data distribution empirically by 

\begin{equation*}
p_{\mathrm{data}}(x) \approx p_{\mathrm{empirical}}(x) = \frac{1}{|D|}\sum_{x_0' \in D}\delta_{x_0'}(x).
\end{equation*}

Applying Bayes with $p_{\mathrm{empirical}}$ in place of $p_{\mathrm{data}}$, we get the \emph{empirical posterior}
\begin{equation}
q_{\mathrm{emp}}\big(x_0 {=} x_0' \mid x_t\big) = \frac{q(x_t \mid x_0')}{\displaystyle\sum_{x_0'' \in D} q(x_t \mid x_0'')}, \qquad x_0' \in D.
\label{eq:empirical-posterior}
\end{equation}
We now evaluate this posterior for each corruption process.

\paragraph{Masked corruption: a hard lookup indicator.}
Let $\mathrm{vis}(x_t)$ denote the set of unmasked positions, and
write $x_t^{\mathrm{vis}}$ for the restriction of $x_t$ to these
positions. Under masked corruption, every visible token agrees
exactly with its clean counterpart, while the probability of masking
a position is independent of its original value. Consequently,
\begin{equation*}
q(x_t \mid x_0')
=
\alpha_t^{\,|\mathrm{vis}(x_t)|}
(1-\alpha_t)^{\,d-|\mathrm{vis}(x_t)|}
\;\ind{x_0'^{\,\mathrm{vis}}=x_t^{\mathrm{vis}}}.
\end{equation*}
For a fixed observation $x_t$, the prefactor is identical for every
candidate $x_0'\in D$ and therefore cancels in the
empirical posterior, Eq.~\eqref{eq:empirical-posterior}. Defining the set of compatible
training samples as
\[
D^{\mathrm{vis}}(x_t)
:=
\big\{
x_0'\in D :
x_0'^{\,\mathrm{vis}}=x_t^{\mathrm{vis}}
\big\},
\]
we obtain, whenever $D^{\mathrm{vis}}(x_t)\neq\varnothing$,
\begin{equation}
q_{\mathrm{emp}}(x_0=x_0' \mid x_t)
=
\frac{\ind{x_0'\in D^{\mathrm{vis}}(x_t)}}
     {|D^{\mathrm{vis}}(x_t)|}.
\label{eq:mdm-global}
\end{equation}
The posterior is thus uniform over training samples that agree with
every visible token and assigns zero probability to all other
samples. This is the global counterpart of the exact lookup of the main text,
Eq.~\eqref{eq:mdm-local-equiv-opt}.

\paragraph{Uniform corruption: a Boltzmann weighting.} The single-token corruption process
$q\big(x_t^{(k)}{=}a \mid x_0^{(k)}{=}b\big) = \alpha_t\,\delta_{a,b} + \frac{1-\alpha_t}{K}$
implies that for the multidimensional expansion we have:
\begin{equation*}
q(x_t \mid x_0') = \Big(\alpha_t + \tfrac{1-\alpha_t}{K}\Big)^{d-\dH(x_t,x_0')}\Big(\tfrac{1-\alpha_t}{K}\Big)^{\dH(x_t,x_0')}
= \Big(\tfrac{1-\alpha_t}{K}\Big)^{d} e^{\beta_t d}\, e^{-\beta_t \dH(x_t,x_0')},
\end{equation*}
with $\beta_t = \log\!\big(1+\frac{\alpha_t K}{1-\alpha_t}\big)$ exactly as
in the Boltzmann kernel of the main text, Eq.~\eqref{eq:optimum}. The prefactor is common to every $x_0' \in D$, so
it cancels in the empirical posterior, Eq.~\eqref{eq:empirical-posterior}, and leaves a Boltzmann
distribution over the training set at inverse temperature $\beta_t$:
\begin{equation}
\begin{aligned}
w(x_0'\mid x_t) &:= q_{\mathrm{emp}}\big(x_0{=}x_0' \mid x_t\big) = \frac{e^{-\beta_t \dH(x_t, x_0')}}{\displaystyle\sum_{x_0''\in D} e^{-\beta_t \dH(x_t, x_0'')}},
\\[4pt]
q_{\mathrm{emp}}\big(x_0^{(k)}{=}c \mid x_t\big) &\phantom{:}= \sum_{x_0' \in D} w(x_0'\mid x_t)\,\ind{x_0'^{(k)}{=}c},
\end{aligned}
\label{eq:udm-global}
\end{equation}
the global counterpart of the unanchored local posterior, Eq.~\eqref{eq:local-equiv-general}, with whole
training samples in place of the patches of $\Ndata$.

 Whereas masked corruption eliminates every sample inconsistent with the visible context, uniform corruption assigns positive posterior weight to every training sample whenever $\alpha_t<1$. The weighting is nevertheless increasingly selective as the noise level decreases:
\[
t\to 0
\quad\Longrightarrow\quad
\alpha_t\to 1
\quad\Longrightarrow\quad
\beta_t\to\infty.
\]
For a fixed input, this limit concentrates the posterior on the set
of training samples at minimum Hamming distance. The next two sections show that both posteriors lead to memorization at sampling time: exactly in the masked case, and with high probability in the uniform case.

\subsection{Masked diffusion memorizes} \label{App: A.3}

The hard consistency constraint of the masked posterior, Eq.~\eqref{eq:mdm-global}, gives a direct memorization
mechanism under exact sequential unmasking. Each newly revealed token
is sampled conditionally on the current visible context, preserving
at least one compatible training sample. The compatible sets are
therefore nested along the reverse trajectory:
\[
\varnothing
\neq D^{\mathrm{vis}}(x_s)
\subseteq D^{\mathrm{vis}}(x_t),
\qquad s<t.
\]
Once all positions are visible, compatibility requires the generated
sample itself to belong to $D$. The same conclusion holds when
multiple tokens are revealed using their exact joint conditional
distribution. It does not generally hold when those tokens are
sampled independently from their marginal posteriors, since
individually admissible token values can be jointly incompatible
with every training sample.

\subsection{Uniform diffusion memorizes under ancestral sampling} \label{App: A.4}

Unlike the masked case, there is no hard constraint that is preserved along the reverse trajectory, so the statement is probabilistic. We first describe the sampler. The argument then has two parts: the last denoising step copies the training sample nearest to the current state (Lemma~\ref{lem:last-step}), and the earlier steps bring the state close to a training sample (Lemma~\ref{lem:factorization}). Proposition~\ref{prop:udm-memorization} combines both. Throughout, we assume that the samples of $D$ are pairwise distinct and denote by
\[
\delta_D := \min_{x_0'\neq x_0''\in D}\dH(x_0',x_0'')
\]
their minimum separation.
\paragraph{Ancestral sampling.} Ancestral sampling generates a sample by running the forward process backwards along a time grid
\[
1=t_N>t_{N-1}>\dots>t_1>t_0=0 .
\]
It starts from $x_{t_N}\sim\mathrm{Unif}(\mathcal V^d)$, which is the forward marginal at $t_N=1$ since $\alpha_1=0$. Then, for $n=N,\dots,1$, it draws $x_{t_{n-1}}$ from a reverse step given $x_{t_n}$. The final state $x_{t_0}$ is the generated sample, which we denote by $x_0$.

\emph{Exact reverse step.} Exact sampling would draw $x_{t_{n-1}}$ from the reverse of the forward process. Conditioning on the clean sample,
\begin{equation}
q\big(x_{t_{n-1}}\mid x_{t_n}\big)=\sum_{x_0\in\mathcal V^d} q\big(x_0\mid x_{t_n}\big)\,q\big(x_{t_{n-1}}\mid x_{t_n},x_0\big),
\label{eq:reverse-general}
\end{equation}
that is, draw a clean sample from the joint posterior $q(x_0\mid x_{t_n})$, and then move towards it. The second factor is known in closed form: given $x_0$ the tokens move independently, $q(x_{t_{n-1}}\mid x_{t_n},x_0)=\prod_k q(x_{t_{n-1}}^{(k)}\mid x_{t_n}^{(k)},x_0^{(k)})$, and each single-token posterior follows from Bayes' rule and the forward kernels. The joint posterior, instead, depends on the data distribution and is not available.

\emph{Ancestral step.} A trained denoiser does not provide this joint posterior so we use the approximation $q(x_0\mid x_{t_n})\approx\prod_{k}q(x_0^{(k)}\mid x_{t_n})\approx\prod_{k}p_\theta(x_0^{(k)}\mid x_{t_n})$. Ancestral sampling~\cite{austin2023structureddenoisingdiffusionmodels} uses the product of these marginals in place of the joint posterior:
\begin{equation}
\begin{aligned}
p^{\mathrm{anc}}\big(x_{t_{n-1}}\mid x_{t_n}\big)
&:=\sum_{x_0\in\mathcal V^d} \Big(\prod_{k=1}^{d}p_\theta\big(x_0^{(k)}\mid x_{t_n}\big)\Big)\,q\big(x_{t_{n-1}}\mid x_{t_n},x_0\big)\\
&\phantom{:}=\prod_{k=1}^{d}\ \sum_{x_0^{(k)}\in\mathcal V}p_\theta\big(x_0^{(k)}\mid x_{t_n}\big)\,q\big(x_{t_{n-1}}^{(k)}\mid x_{t_n}^{(k)},x_0^{(k)}\big).
\end{aligned}
\label{eq:ancestral-general}
\end{equation}

At the empirical optimum, the denoiser is the marginal of the Boltzmann posterior,
\[
p_\theta^{\star}\big(x_0^{(k)}\mid x_{t_n}\big)=q_{\mathrm{emp}}\big(x_0^{(k)}\mid x_{t_n}\big)=\sum_{x_0'\in D}w(x_0'\mid x_{t_n})\,\ind{x_0'^{(k)}=x_0^{(k)}} .
\]
Substituting it and summing over $x_0^{(k)}$, the indicator keeps only the term $x_0^{(k)}=x_0'^{(k)}$:

\begin{equation}
p^{\mathrm{anc}}\big(x_{t_{n-1}}\mid x_{t_n}\big)
=\prod_{k=1}^{d}\ \sum_{x_0'\in D} w(x_0'\mid x_{t_n})\;
q\big(x_{t_{n-1}}^{(k)}\mid x_{t_n}^{(k)},x_0'^{(k)}\big).
\label{eq:ancestral}
\end{equation}

\paragraph{Last step.} By the previous display, the generated sample $x_0$ depends on the trajectory only through $x_{t_1}$. We therefore start from the end: at small noise the weights $w(x_0'\mid x_{t_1})$ concentrate on the training samples nearest to $x_{t_1}$, and if $x_{t_1}$ is close enough to one of them, the last step copies it.

\begin{lemma}[The last step copies the nearest training sample]
\label{lem:last-step}
Let $D^\star(x_t)$ be the set of training samples at minimum Hamming distance from $x_t$.
\begin{enumerate}
\item[(i)] For fixed $x_t$, every position $k$ and every $x_0^{(k)}\in\mathcal V$,
\[
\lim_{\beta_t\to\infty} q_{\mathrm{emp}}\big(x_0^{(k)}\mid x_t\big)
= \frac{1}{|D^\star(x_t)|}\sum_{x_0'\in D^\star(x_t)} \ind{x_0'^{(k)}=x_0^{(k)}}.
\]
\item[(ii)] If some $x^\star\in D$ satisfies $r:=\dH(x_{t_1},x^\star)<\delta_D/2$, then $D^\star(x_{t_1})=\{x^\star\}$ and the last ancestral step satisfies
\[
\mathbb P\big(x_0\neq x^\star\mid x_{t_1}\big)\;\le\; d\,(|D|-1)\,e^{-\beta_{t_1}(\delta_D-2r)}.
\]
\end{enumerate}
\end{lemma}

\begin{proof}
Let $\Delta(x_0'):=\dH(x_t,x_0')-\min_{x_0''\in D}\dH(x_t,x_0'')\ge 0$, which vanishes exactly on $D^\star(x_t)$. Dividing numerator and denominator of the Boltzmann weights, Eq.~\eqref{eq:udm-global}, by the largest exponential, $e^{-\beta_t\min_{x_0''}\dH(x_t,x_0'')}$,
\[
w(x_0'\mid x_t)=\frac{e^{-\beta_t\Delta(x_0')}}{\sum_{x_0''\in D}e^{-\beta_t\Delta(x_0'')}}
\;\xrightarrow[\beta_t\to\infty]{}\;
\frac{\ind{x_0'\in D^\star(x_t)}}{|D^\star(x_t)|},
\]
since $e^{-\beta_t\Delta}$ tends to $1$ when $\Delta=0$ and to $0$ when $\Delta>0$. Part (i) follows because $q_{\mathrm{emp}}(x_0^{(k)}\mid x_t)=\sum_{x_0'}w(x_0'\mid x_t)\,\ind{x_0'^{(k)}=x_0^{(k)}}$ is a finite sum of these weights.

For (ii), let us first see that every other training sample is strictly farther from $x_{t_1}$ than $x^\star$. Take $x_0'\in D$ with $x_0'\neq x^\star$. By definition of $\delta_D$ and the triangle inequality,
\[
\delta_D\;\le\;\dH(x^\star,x_0')\;\le\;\dH(x^\star,x_{t_1})+\dH(x_{t_1},x_0') .
\]
Rearranging,
\[
\dH(x_{t_1},x_0')\;\ge\;\delta_D-\dH(x_{t_1},x^\star) .
\]
Subtracting $\dH(x_{t_1},x^\star)=r$ on both sides,
\[
\dH(x_{t_1},x_0')-\dH(x_{t_1},x^\star)\;\ge\;\delta_D-2r\;>\;0 ,
\]
where the last inequality is the assumption $r<\delta_D/2$. Hence $x^\star$ is the unique nearest sample, $D^\star(x_{t_1})=\{x^\star\}$, and
\[
\Delta(x^\star)=0,
\qquad
\Delta(x_0')\;\ge\;\delta_D-2r\quad\text{for every } x_0'\neq x^\star .
\]

Next, we bound the weight that is not on $x^\star$. From the expression of $w$ above, evaluated at $x_{t_1}$,
\[
1-w(x^\star\mid x_{t_1})
=\frac{\sum_{x_0'\neq x^\star}e^{-\beta_{t_1}\Delta(x_0')}}{1+\sum_{x_0'\neq x^\star}e^{-\beta_{t_1}\Delta(x_0')}}
\;\le\;\sum_{x_0'\neq x^\star}e^{-\beta_{t_1}\Delta(x_0')}
\;\le\;(|D|-1)\,e^{-\beta_{t_1}(\delta_D-2r)} .
\]

Finally, we pass from the weights to the generated sample. The last step draws each token from $q_{\mathrm{emp}}(x_0^{(k)}\mid x_{t_1})$, and $x^\star$ contributes its weight to the value $x^{\star(k)}$, so
\[
\mathbb P\big(x_0^{(k)}\neq x^{\star(k)}\mid x_{t_1}\big)
=1-q_{\mathrm{emp}}\big(x_0^{(k)}{=}x^{\star(k)}\mid x_{t_1}\big)
\;\le\;1-w(x^\star\mid x_{t_1}) .
\]
The sample differs from $x^\star$ only if at least one of its $d$ tokens does, so a union bound gives
\[
\mathbb P\big(x_0\neq x^\star\mid x_{t_1}\big)
\;\le\;d\,\big(1-w(x^\star\mid x_{t_1})\big)
\;\le\;d\,(|D|-1)\,e^{-\beta_{t_1}(\delta_D-2r)} .
\]
\end{proof}

\paragraph{Middle steps.} The last-step lemma requires $x_{t_1}$ to be close to a training sample. Let $p^{\mathrm{anc}}(x_{t_n})$ denote the law of the state of the sampler at time $t_n$. It has no closed form, so we compare it with the forward marginal $q_{\mathrm{emp}}(x_{t_n})$, the law of a training sample noised up to time $t_n$. The two laws coincide at $t_N=1$, where both are uniform since $\alpha_1=0$. From there, the forward marginals are propagated by the exact step $q_{\mathrm{emp}}(x_{t_{n-1}}\mid x_{t_n})$, and the laws of the sampler by the ancestral step $p^{\mathrm{anc}}(x_{t_{n-1}}\mid x_{t_n})$. If the two steps stay close at every time, then
\[
p^{\mathrm{anc}}(x_{t_1})\;\approx\;q_{\mathrm{emp}}(x_{t_1}),
\]
and the requirement holds once $t_1$ is small, since a training sample noised up to a small time has few altered tokens. The next lemma shows that the two steps are indeed close: they differ only in the order of the sum over $x_0'$ and the product over $k$, which matters only when two or more tokens move in the same step.

For two probability distributions $P(x)$ and $Q(x)$ over $x\in\mathcal V^d$, we measure their distance by the total variation \cite{levin2017markov}:

\[
\mathrm{TV}(P,Q):=\frac12\sum_{x\in\mathcal V^d}\big|P(x)-Q(x)\big|=\max_{A\subseteq\mathcal V^d}\big|P(A)-Q(A)\big| .
\]
The second form is the property we use to find the following result:

\begin{lemma}[Factorization error]
\label{lem:factorization}
For $2\le n\le N$, define
\[
h_n:=\frac{\alpha_{t_{n-1}}-\alpha_{t_n}}{\alpha_{t_{n-1}}\,(1-\alpha_{t_n})}\quad(n<N),
\qquad
h_N:=\alpha_{t_{N-1}} .
\]
Then, for every $x_{t_n}\in\mathcal V^d$,
\[
\mathrm{TV}\Big(p^{\mathrm{anc}}\big(x_{t_{n-1}}\mid x_{t_n}\big),\ q_{\mathrm{emp}}\big(x_{t_{n-1}}\mid x_{t_n}\big)\Big)\;\le\;\frac{(d\,h_n)^2}{2},
\]
where the total variation is taken over $x_{t_{n-1}}\in\mathcal V^d$.
\end{lemma}

The idea of the proof is the following. In $q_{\mathrm{emp}}$ all tokens follow one common training sample, whereas in $p^{\mathrm{anc}}$ each token follows its own. This would make a large difference if all tokens moved. In one step, however, each token moves only with probability about $h_n$, and a token that does not move is unaffected by the training sample it follows. The two kernels can therefore only disagree when at least two tokens move, which has probability at most $(d\,h_n)^2/2$.

\begin{proof}
Fix $n$ and $x_{t_n}$. Under $p_{\mathrm{empirical}}$, the posterior of the clean sample is supported on $D$, with $q_{\mathrm{emp}}(x_0{=}x_0'\mid x_{t_n})=w(x_0'\mid x_{t_n})$. Given the clean sample, the tokens move independently. The two kernels we compare are therefore
\begin{align*}
q_{\mathrm{emp}}\big(x_{t_{n-1}}\mid x_{t_n}\big)&=\sum_{x_0'\in D}w(x_0'\mid x_{t_n})\,\prod_{k=1}^{d}q\big(x_{t_{n-1}}^{(k)}\mid x_{t_n}^{(k)},x_0'^{(k)}\big),\\
p^{\mathrm{anc}}\big(x_{t_{n-1}}\mid x_{t_n}\big)&=\prod_{k=1}^{d}\ \sum_{x_0'\in D}w(x_0'\mid x_{t_n})\,q\big(x_{t_{n-1}}^{(k)}\mid x_{t_n}^{(k)},x_0'^{(k)}\big),
\end{align*}
where the second line is Eq.~\eqref{eq:ancestral}. They differ only in the order of the sum over $x_0'$ and the product over $k$.

A token stays with probability at least $1-h_n$. Let first $n<N$, and let $\lambda:=\alpha_{t_n}/\alpha_{t_{n-1}}$ be the probability that the forward process keeps a token between $t_{n-1}$ and $t_n$. By Bayes' rule,
\[
q\big(x_{t_{n-1}}^{(k)}\mid x_{t_n}^{(k)},x_0^{(k)}\big)=\frac{q\big(x_{t_n}^{(k)}\mid x_{t_{n-1}}^{(k)}\big)\,q\big(x_{t_{n-1}}^{(k)}\mid x_0^{(k)}\big)}{q\big(x_{t_n}^{(k)}\mid x_0^{(k)}\big)} .
\]
For $x_{t_{n-1}}^{(k)}=x_{t_n}^{(k)}$, the first factor of the numerator is $\lambda+(1-\lambda)/K$, so
\[
q\big(x_{t_{n-1}}^{(k)}{=}x_{t_n}^{(k)}\mid x_{t_n}^{(k)},x_0^{(k)}\big)
=\Big(\lambda+\frac{1-\lambda}{K}\Big)\,
\frac{q\big(x_{t_{n-1}}^{(k)}{=}x_{t_n}^{(k)}\mid x_0^{(k)}\big)}{q\big(x_{t_n}^{(k)}\mid x_0^{(k)}\big)} .
\]
If $x_{t_n}^{(k)}\neq x_0^{(k)}$, the ratio equals $(1-\alpha_{t_{n-1}})/(1-\alpha_{t_n})$, and therefore
\[
q\big(x_{t_{n-1}}^{(k)}{=}x_{t_n}^{(k)}\mid x_{t_n}^{(k)},x_0^{(k)}\big)
\;\ge\;\lambda\,\frac{1-\alpha_{t_{n-1}}}{1-\alpha_{t_n}}
\;=\;\frac{\alpha_{t_n}(1-\alpha_{t_{n-1}})}{\alpha_{t_{n-1}}(1-\alpha_{t_n})}
\;=\;1-h_n .
\]
If $x_{t_n}^{(k)}=x_0^{(k)}$, the ratio equals $\frac{\alpha_{t_{n-1}}+(1-\alpha_{t_{n-1}})/K}{\alpha_{t_n}+(1-\alpha_{t_n})/K}\ge1$, since $\alpha_{t_{n-1}}\ge\alpha_{t_n}$. The probability of staying is then at least $\lambda\ge1-h_n$.

By the bound above, for every $x_0'\in D$ we can write
\[
q\big(x_{t_{n-1}}^{(k)}\mid x_{t_n}^{(k)},x_0'^{(k)}\big)=(1-h_n)\,\delta_{x_{t_n}^{(k)}}\big(x_{t_{n-1}}^{(k)}\big)+h_n\,\nu^{(k)}_{x_0'}\big(x_{t_{n-1}}^{(k)}\big),
\]
where
\[
\nu^{(k)}_{x_0'}\big(x_{t_{n-1}}^{(k)}\big):=\frac{q\big(x_{t_{n-1}}^{(k)}\mid x_{t_n}^{(k)},x_0'^{(k)}\big)-(1-h_n)\,\delta_{x_{t_n}^{(k)}}\big(x_{t_{n-1}}^{(k)}\big)}{h_n}
\]
is non-negative and sums to one, so it is a distribution over $x_{t_{n-1}}^{(k)}$. A token thus stays at $x_{t_n}^{(k)}$ with probability $1-h_n$, and otherwise moves according to $\nu^{(k)}_{x_0'}$. The probability of moving, $h_n$, is the same for every $x_0'$: only where the token moves depends on $x_0'$.

We insert this decomposition into both kernels and expand the product over $k$. Each term of the expansion picks, for every token, either the first summand (the token stays) or the second (the token moves). We index the terms by the set $S\subseteq\{1,\dots,d\}$ of moving tokens, and regard $S$ as random, with each token in $S$ independently with probability $h_n$, so that $\mathbb P(S)=h_n^{|S|}(1-h_n)^{d-|S|}$. In $q_{\mathrm{emp}}$ this gives
\[
q_{\mathrm{emp}}\big(x_{t_{n-1}}\mid x_{t_n}\big)
=\sum_{S}\mathbb P(S)\prod_{k\notin S}\delta_{x_{t_n}^{(k)}}\big(x_{t_{n-1}}^{(k)}\big)\,
\sum_{x_0'\in D}w(x_0'\mid x_{t_n})\prod_{k\in S}\nu^{(k)}_{x_0'}\big(x_{t_{n-1}}^{(k)}\big).
\]
In $p^{\mathrm{anc}}$ the sum over $x_0'$ stands inside each factor. Since the weights $w(x_0'\mid x_{t_n})$ sum to one, it only acts on the second summand, and each factor equals $(1-h_n)\,\delta_{x_{t_n}^{(k)}}+h_n\sum_{x_0'}w(x_0'\mid x_{t_n})\,\nu^{(k)}_{x_0'}$. Expanding in the same way,
\[
p^{\mathrm{anc}}\big(x_{t_{n-1}}\mid x_{t_n}\big)
=\sum_{S}\mathbb P(S)\prod_{k\notin S}\delta_{x_{t_n}^{(k)}}\big(x_{t_{n-1}}^{(k)}\big)\,
\prod_{k\in S}\ \sum_{x_0'\in D}w(x_0'\mid x_{t_n})\,\nu^{(k)}_{x_0'}\big(x_{t_{n-1}}^{(k)}\big).
\]
Both kernels draw $S$ in the same way and keep the tokens outside $S$ at $x_{t_n}$. They differ only in the last factor, which says where the tokens in $S$ go: in $q_{\mathrm{emp}}$ they follow one common $x_0'$, and in $p^{\mathrm{anc}}$ each follows its own. If $S$ is empty, both last factors equal one. If $S=\{k\}$, both equal $\sum_{x_0'}w(x_0'\mid x_{t_n})\,\nu^{(k)}_{x_0'}\big(x_{t_{n-1}}^{(k)}\big)$, since with a single moving token a common $x_0'$ and an individual one give the same law. The two kernels can therefore only differ when $|S|\ge2$.

\emph{Bound.} Both kernels are mixtures over $S$ with the same weights $\mathbb P(S)$. Applying the triangle inequality inside the sum that defines $\mathrm{TV}$, the total variation between two such mixtures is at most the $\mathbb P(S)$-weighted sum of the total variations between their components. Components with $|S|\le1$ coincide, and the others are at total variation at most one, so
\[
\mathrm{TV}\Big(p^{\mathrm{anc}}\big(x_{t_{n-1}}\mid x_{t_n}\big),\ q_{\mathrm{emp}}\big(x_{t_{n-1}}\mid x_{t_n}\big)\Big)
\;\le\;\mathbb P\big(|S|\ge 2\big).
\]
At least two tokens move only if some pair of tokens moves. There are $\binom d2$ pairs, and each moves with probability $h_n^2$, so
\[
\mathbb P\big(|S|\ge2\big)\;\le\;\binom d2\, h_n^2\;\le\;\frac{(d\,h_n)^2}{2}.
\]

\emph{The first step, $n=N$.} Since $\alpha_{t_N}=0$, the forward process resamples every token uniformly between $t_{N-1}$ and $t_N$, so $x_{t_N}$ carries no information and
\[
q\big(x_{t_{N-1}}^{(k)}\mid x_{t_N}^{(k)},x_0^{(k)}\big)=q\big(x_{t_{N-1}}^{(k)}\mid x_0^{(k)}\big)=(1-\alpha_{t_{N-1}})\,\frac1K+\alpha_{t_{N-1}}\,\delta_{x_0^{(k)}}\big(x_{t_{N-1}}^{(k)}\big).
\]
This is the same decomposition with $h_N=\alpha_{t_{N-1}}$, the uniform distribution in place of $\delta_{x_{t_N}^{(k)}}$, and $\nu^{(k)}_{x_0'}=\delta_{x_0'^{(k)}}$: a token is resampled uniformly with probability $1-h_N$, and jumps to $x_0'^{(k)}$ otherwise. The tokens outside $S$ are resampled uniformly in both kernels, and the rest of the argument is unchanged.
\end{proof}

We now combine the two lemmas. The factorization-error lemma transfers the law of $x_{t_1}$ from the exact reverse process to the ancestral sampler, and the last-step lemma handles the last step.

\begin{proposition}[Uniform diffusion memorizes]
\label{prop:udm-memorization}
Let $x_0$ be generated by ancestral sampling at the empirical optimum, with the step $p^{\mathrm{anc}}(x_{t_{n-1}}\mid x_{t_n})$ of Eq.~\eqref{eq:ancestral}. Let $h_n$ be as in Lemma~\ref{lem:factorization}, and let $\rho:=(1-\alpha_{t_1})\frac{K-1}{K}$. Then
\begin{equation}
\mathbb P\big(x_0\notin D\big)
\;\le\;
\frac{d^2}{2}\sum_{n=2}^{N} h_n^2
\;+\;
\frac{4d\,\rho}{\delta_D}
\;+\;
d\,(|D|-1)\,e^{-\beta_{t_1}\delta_D/2} .
\label{eq:udm-memorization-bound}
\end{equation}
For a continuous schedule $\alpha_t$, the right-hand side can be made arbitrarily small by refining the time grid, so the sampler returns a training sample with probability arbitrarily close to one.
\end{proposition}

The proof follows the chain
\begin{enumerate}
\item[(a)] $p^{\mathrm{anc}}(x_{t_1})\approx q_{\mathrm{emp}}(x_{t_1})$: the state of the sampler at time $t_1$ is distributed almost like a training sample noised up to time $t_1$.
\item[(b)] such a state differs from some $x_0'\in D$ in only a few tokens.
\item[(c)] from such a state, the last step returns $x_0'$.
\end{enumerate}
Each step fails with small probability, and the three failure probabilities are the three terms of Eq.~\eqref{eq:udm-memorization-bound}.

\begin{proof}
Let $p^{\mathrm{anc}}(x_{t_n})$ be the law of the state of the sampler at time $t_n$, and $q_{\mathrm{emp}}(x_{t_n})$ the law of a training sample $x_0'\sim\mathrm{Unif}(D)$ noised up to time $t_n$ by the forward process. Write
\[
\varepsilon_n:=\mathrm{TV}\big(p^{\mathrm{anc}}(x_{t_n}),\,q_{\mathrm{emp}}(x_{t_n})\big).
\]

\emph{Step (a).} At $t_N=1$ both laws are uniform: the sampler is initialized there, and the forward marginal is uniform because $\alpha_1=0$. Hence $\varepsilon_N=0$. From one time to the next, each law is propagated by its own step,
\begin{align*}
p^{\mathrm{anc}}(x_{t_{n-1}})&=\sum_{x_{t_n}}p^{\mathrm{anc}}(x_{t_n})\,p^{\mathrm{anc}}\big(x_{t_{n-1}}\mid x_{t_n}\big),\\
q_{\mathrm{emp}}(x_{t_{n-1}})&=\sum_{x_{t_n}}q_{\mathrm{emp}}(x_{t_n})\,q_{\mathrm{emp}}\big(x_{t_{n-1}}\mid x_{t_n}\big),
\end{align*}
where the second line holds because $q_{\mathrm{emp}}(x_{t_{n-1}}\mid x_{t_n})$ is the conditional law of $x_{t_{n-1}}$ given $x_{t_n}$ under the forward process. To compare them, we insert the intermediate law
\[
\mu(x_{t_{n-1}}):=\sum_{x_{t_n}}p^{\mathrm{anc}}(x_{t_n})\,q_{\mathrm{emp}}\big(x_{t_{n-1}}\mid x_{t_n}\big),
\]
which applies the exact step to the state of the sampler. Then $p^{\mathrm{anc}}(x_{t_{n-1}})$ and $\mu$ share the law of $x_{t_n}$ but use different steps, while $\mu$ and $q_{\mathrm{emp}}(x_{t_{n-1}})$ use the same step on different laws of $x_{t_n}$. In both cases the triangle inequality inside the sum that defines $\mathrm{TV}$ gives a bound. For the first pair,
\begin{align*}
\mathrm{TV}\big(p^{\mathrm{anc}}(x_{t_{n-1}}),\,\mu\big)
&=\frac12\sum_{x_{t_{n-1}}}\Big|\sum_{x_{t_n}}p^{\mathrm{anc}}(x_{t_n})\Big(p^{\mathrm{anc}}\big(x_{t_{n-1}}\mid x_{t_n}\big)-q_{\mathrm{emp}}\big(x_{t_{n-1}}\mid x_{t_n}\big)\Big)\Big|\\
&\le\sum_{x_{t_n}}p^{\mathrm{anc}}(x_{t_n})\,\mathrm{TV}\Big(p^{\mathrm{anc}}\big(x_{t_{n-1}}\mid x_{t_n}\big),\,q_{\mathrm{emp}}\big(x_{t_{n-1}}\mid x_{t_n}\big)\Big)
\;\le\;\frac{(d\,h_n)^2}{2},
\end{align*}
where the last inequality is the factorization-error lemma (Lemma~\ref{lem:factorization}), which holds for every $x_{t_n}$. For the second pair, using that $q_{\mathrm{emp}}(x_{t_{n-1}}\mid x_{t_n})$ sums to one over $x_{t_{n-1}}$,
\[
\mathrm{TV}\big(\mu,\,q_{\mathrm{emp}}(x_{t_{n-1}})\big)
=\frac12\sum_{x_{t_{n-1}}}\Big|\sum_{x_{t_n}}\big(p^{\mathrm{anc}}(x_{t_n})-q_{\mathrm{emp}}(x_{t_n})\big)\,q_{\mathrm{emp}}\big(x_{t_{n-1}}\mid x_{t_n}\big)\Big|
\;\le\;\varepsilon_n :
\]
applying the same step to two laws cannot increase their distance. Adding the two bounds,
\[
\varepsilon_{n-1}\;\le\;\varepsilon_n+\frac{(d\,h_n)^2}{2},
\]
and summing from $\varepsilon_N=0$ down to $n=2$,
\[
\varepsilon_1\;\le\;\frac{d^2}{2}\sum_{n=2}^{N}h_n^2 .
\]

The last step uses the same kernel whatever the law of $x_{t_1}$. Writing $f(x_{t_1})\in[0,1]$ for the probability that it returns a sample outside $D$ from the state $x_{t_1}$,
\[
\mathbb P\big(x_0\notin D\big)=\sum_{x_{t_1}}p^{\mathrm{anc}}(x_{t_1})\,f(x_{t_1})
\;\le\;\sum_{x_{t_1}}q_{\mathrm{emp}}(x_{t_1})\,f(x_{t_1})+\varepsilon_1 ,
\]
since $\sum_x\big(P(x)-Q(x)\big)f(x)$ is at most the total mass of the positive part of $P-Q$, which is $\mathrm{TV}(P,Q)$. It remains to bound the probability of leaving $D$ when $x_{t_1}\sim q_{\mathrm{emp}}(x_{t_1})$.

\emph{Step (b).} Under $q_{\mathrm{emp}}(x_{t_1})$, the state $x_{t_1}$ is a training sample $x_0'$ in which every token has been altered independently with probability $\rho$: a token is resampled with probability $1-\alpha_{t_1}$, and the new value differs from the old one with probability $\frac{K-1}{K}$. The number of altered tokens, $m:=\dH(x_{t_1},x_0')$, therefore has mean $d\rho$, and by Markov's inequality
\[
\mathbb P\Big(m>\frac{\delta_D}{4}\Big)\;\le\;\frac{4d\rho}{\delta_D}.
\]

\emph{Step (c).} On the complementary event, $r:=\dH(x_{t_1},x_0')\le\delta_D/4<\delta_D/2$. Part (ii) of the last-step lemma (Lemma~\ref{lem:last-step}) states that, whenever $\dH(x_{t_1},x^\star)=r<\delta_D/2$ for some $x^\star\in D$, the last step returns $x^\star$ except with probability $d\,(|D|-1)\,e^{-\beta_{t_1}(\delta_D-2r)}$. With $x^\star=x_0'$ and $\delta_D-2r\ge\delta_D/2$, the last step returns $x_0'\in D$ except with probability
\[
d\,(|D|-1)\,e^{-\beta_{t_1}\delta_D/2}.
\]
We now put the three steps together. Under $q_{\mathrm{emp}}(x_{t_1})$, the state $x_{t_1}$ is obtained by noising a training sample $x_0'\sim\mathrm{Unif}(D)$, and we split according to the event of step (b):
\[
\sum_{x_{t_1}}q_{\mathrm{emp}}(x_{t_1})\,f(x_{t_1})
=\mathbb E\Big[f(x_{t_1})\,\ind{m>\delta_D/4}\Big]+\mathbb E\Big[f(x_{t_1})\,\ind{m\le\delta_D/4}\Big],
\]
where the expectations are over $x_0'$ and $x_{t_1}$. Inserting both bounds and the bound on $\varepsilon_1$ from step (a),
\begin{align*}
\mathbb P\big(x_0\notin D\big)
&\le\varepsilon_1+\sum_{x_{t_1}}q_{\mathrm{emp}}(x_{t_1})\,f(x_{t_1})\\
&\le\frac{d^2}{2}\sum_{n=2}^{N}h_n^2+\frac{4d\rho}{\delta_D}+d\,(|D|-1)\,e^{-\beta_{t_1}\delta_D/2},
\end{align*}
which is Eq.~\eqref{eq:udm-memorization-bound}.

\emph{Refining the grid.} We make each term small in turn. First choose $t_1$ close to $0$: then $\rho\to0$ and $\beta_{t_1}\to\infty$, so the last two terms vanish. Next choose $t_{N-1}$ close to $1$, so that $h_N=\alpha_{t_{N-1}}$ is small. With $t_1$ and $t_{N-1}$ fixed, consider $2\le n\le N-1$. Then $\alpha_{t_{n-1}}\ge\alpha_{t_{N-1}}$ and $1-\alpha_{t_n}\ge1-\alpha_{t_1}$, so
\[
h_n\;\le\;\frac{\alpha_{t_{n-1}}-\alpha_{t_n}}{\alpha_{t_{N-1}}(1-\alpha_{t_1})}.
\]
The increments $\alpha_{t_{n-1}}-\alpha_{t_n}$ are non-negative and sum to at most one, hence
\[
\sum_{n=2}^{N-1}h_n^2\;\le\;\frac{\max_{n}\,(\alpha_{t_{n-1}}-\alpha_{t_n})}{\alpha_{t_{N-1}}^2(1-\alpha_{t_1})^2},
\]
which vanishes as the grid between $t_1$ and $t_{N-1}$ is refined, by continuity of $\alpha_t$.
\end{proof}

The first term of the bound in Eq.~\eqref{eq:udm-memorization-bound}, the discretization error, is pessimistic: it only uses that two tokens rarely move in the same step. In practice the weights $w(x_0'\mid x_{t_n})$ concentrate on a single training sample well before the last step, and from then on every token is drawn towards that same sample.

\paragraph{Theoretical summary.} The table below collects the differences between the two corruption processes.
\begin{center}
\small
\setlength{\tabcolsep}{5pt}
\renewcommand{\arraystretch}{1.1}
\begin{tabular}{@{}lcc@{}}
\toprule
& \textbf{Uniform} & \textbf{Masked} \\
\midrule
Kernel
& $e^{-\beta_t \dH(x_t,x_0')}$
& $\ind{x_0'^{\,\mathrm{vis}}=x_t^{\mathrm{vis}}}$ \\
Support
& $D$ ($\alpha_t<1$)
& $D^{\mathrm{vis}}(x_t)\subseteq D$ \\
Weighting
& Boltzmann
& Uniform on compatible samples \\
Selection
& Sharpens as $t\to0$
& Exact visible-token matching \\
Memorization
& Almost exact, depends on discretisation
& Exact (nested compatible sets) \\
\bottomrule
\end{tabular}
\end{center}

\subsection{Numerical verification of memorization in the uniform case} \label{App: A.5}

We run ancestral sampling with the Boltzmann posterior of Eq.~\eqref{eq:udm-global} and measure how close each generated sample is to the training set. Table~\ref{tab:memo-ancestral} reports $200$ independent generations at
each of $K \in \{2,4,8\}$ on CIFAR-10 with $|D|{=}5000$ training samples, $600$ finished samples in total.

\begin{table}[ht!]
\centering
\begin{tabular}{lcc}
\toprule
$K$ & Mean match & Worst run \\
\midrule
2 & 99.76\% & 99.32\% \\
4 & 99.63\% & 99.02\% \\
8 & 99.54\% & 98.83\% \\
\bottomrule
\end{tabular}

\vspace{8pt}
\caption{Fraction of tokens shared between a generated sample
and its single nearest training sample: mean and minimum over $200$ independent
generations per $K$ (CIFAR-10, $|D|{=}5000$, $40$ schedule steps, samples
initialized from pure noise).}
\label{tab:memo-ancestral}
\end{table}

Every run finishes with more than $98.8\%$ of its tokens equal to those of the closest training sample, and the mean agreement exceeds $99.5\%$ for every $K$. With only $40$ steps the grid is far coarser than the memorization bound of Proposition~\ref{prop:udm-memorization} requires, so most samples remain a few tokens away from perfect memorization. This difference is irrelevant for our purposes: a match of more than $90\%$ already shows clear memorization behaviour.
\newpage
\section{Theoretical proofs for Uniform Diffusion}
\label{app:theory-udm}

\subsection{Deriving the UDM Locality optimum}
We start by reframing the loss for our purposes. For a fixed $t$, the uniform diffusion loss is

\begin{equation*}
\mathcal{L}(\theta) = \mathbb{E}_{t \sim U(0,1)} \left[ \mathcal{L}_t(\theta) \right],
\end{equation*}

\begin{equation*}
\mathcal{L}_t(\theta) = \mathbb{E}_{q(x_0)}\left[\mathbb{E}_{q(x_t\mid x_0)}\left[\sum_{k=1}^d -\log p_\theta\big(x_0^{(k)}\mid x_t\big)\right]\right],
\end{equation*}

Using $q(x_0)\,q(x_t\mid x_0)=q(x_t)\,q(x_0\mid x_t)$ and separating $x_0=(x_0^{(k)}, x_0^{/k})$ we marginalise over $x_0^{/k}$:

\begin{equation}
\mathcal{L}_t(\theta) = \mathbb{E}_{q(x_t)}\left[\sum_{k=1}^d \mathbb{E}_{q(x_0^{(k)}\mid x_t)}\Big[-\log p_\theta\big(x_0^{(k)}\mid x_t\big)\Big]\right].
\label{eq:udm-loss-marginal}
\end{equation}

With this formulation, we now proceed to prove Proposition~\ref{prop:locality}.

\noindent\textbf{Proof of Proposition~\ref{prop:locality} (Local Machine).}
Restrict $p_\theta$ to a class of models observing only a local patch $\Ntk\subset x_t$, so that $p_\theta(x_0^{(k)}\mid x_t)\equiv p_\theta(x_0^{(k)}\mid \Ntk)$.
\begin{proof}
Decomposing $x_t$ into $\Ntk$ and the non-local remainder $x_{t,\backslash \Ntk}$ gives 

\[q(x_t)\,q(x_0^{(k)}\mid x_t) = q(x_0^{(k)},\Ntk,x_{t,\backslash \Ntk}).\] 

Since $-\log p_\theta(x_0^{(k)}\mid \Ntk)$ does not depend on $x_{t,\backslash \Ntk}$, marginalizing it out of \eqref{eq:udm-loss-marginal} term by term gives $\sum_{x_{t,\backslash \Ntk}} q(x_0^{(k)},\Ntk,x_{t,\backslash \Ntk}) = q(\Ntk)\,q(x_0^{(k)}\mid \Ntk)$, so
\begin{equation*}
\mathcal{L}_t(\theta) = \sum_{k=1}^d \mathbb{E}_{q(\Ntk)}\Big[\mathbb{E}_{q(x_0^{(k)}\mid \Ntk)}\big[-\log p_\theta\big(x_0^{(k)}\mid \Ntk\big)\big]\Big].
\end{equation*}

Now the loss is reframed identically to the standard UDM Cross Entropy loss but for a local window. The minimizer of the Cross Entropy, by Gibbs' inequality, is $p_\theta^*(x_0^{(k)}|\Ntk)=q(x_0^{(k)}|\Ntk)$.
\end{proof}

\subsection{Equivariant models: definitions and setup}

In this section we will consider the following inductive bias on our model, restricting its function class. For some group $G$, $p_\theta$ is said to be \emph{equivariant under $G$} if it follows that:

\begin{equation*}
p_\theta\big(x_0^{(k)} \mid g\cdot x_t\big) = p_\theta\big(x_0^{(g^{-1}\cdot k)} \mid  x_t\big), \qquad \forall\, g \in G,\ \forall\, k,\ \forall\, x_t,\ \forall\, x_0^{(k)}\in\mathcal V.
\end{equation*}

This is the standard definition of equivariance. However, by a simple reindexing, it can be rewritten in a more convenient \textit{invariance} form. Letting $k' = g^{-1}\!\cdot k$, we obtain:

\begin{equation*}
p_\theta\big(x_0^{(g\cdot k')} \mid g\cdot x_t\big) = p_\theta\big(x_0^{(k')} \mid  x_t\big), \qquad \forall\, g \in G,\ \forall\, k',\ \forall\, x_t,\ \forall\, x_0^{(k')}\in\mathcal V.
\end{equation*}

We will use the words \textit{equivariant} and \textit{invariant} interchangeably depending on the context. 

The action we will consider in particular is the following exhaustive token action on coordinates, of which translation is a particular case:

\begin{definition*}[Coordinate action of $G$]
A \emph{token} $x_t^{(k)}$ is a pair $(s,k)\in\mathcal V\times\{1,\dots,d\}$, read as ``value $s$ sitting at coordinate $k$.'' Let $G$ act on coordinates through a group homomorphism $g\mapsto\sigma_g$ into the symmetric group on $\{1,\dots,d\}$, and write $g\cdot k := \sigma_g(k)$, so that
\begin{equation*}
e\cdot k = k, \qquad g\cdot(g'\cdot k) = (gg')\cdot k, \qquad \forall\, g,g'\in G,\ k\in\{1,\dots,d\}.
\end{equation*}
This action acts on tokens by leaving the value untouched,
\begin{equation*}
g\cdot(s,k) := (s,\ g\cdot k), \qquad s\in\mathcal V,\ k\in\{1,\dots,d\},
\end{equation*}
which for each fixed $g$ is a bijection of $\mathcal V\times\{1,\dots,d\}$ (identity on the $\mathcal V$-factor, the bijection $\sigma_g$ on the coordinate factor), and in particular injective.
\end{definition*}

The action of $G$ extends from coordinates to sequences $x\in\mathcal V^d$ by relocating each token: the token $(x^{(k)},k)$ is sent to $(x^{(k)}, g\cdot k)$, i.e.\ the value $x^{(k)}$ now sits at coordinate $g\cdot k$ instead of $k$. Writing this relocation as a new sequence $g\cdot x$ gives
\begin{equation*}
(g\cdot x)^{(j)} := x^{(g^{-1}\cdot j)}, \qquad j \in \{1,\dots,d\},
\end{equation*}
from which, setting $j = g\cdot k$,
\begin{equation}
(g\cdot x)^{(g\cdot k)} = x^{(k)} \qquad \forall\, x \in \mathcal V^d.
\label{eq:seq-action-cancel}
\end{equation}
As before, we restrict to single-orbit actions: $G\cdot k = \{1,\dots,d\}$ for every $k$.

\begin{remark*}
Our equivariance results hold for any coordinate action of the form defined above. We only require that $g\cdot$ induces a bijection over coordinates. Later, when deriving our dataset amplification result, we will restrict attention to the translation group $\mathcal{T}$, whose action preserves the spatial structure of patches and samples.
\end{remark*}

With the $G$-action now defined, we can show that the cross-entropy objective used in Uniform Diffusion and Masked Diffusion is itself equivariant whenever the denoising architecture is invariant under $G$.

Write the summed per-token cross-entropy loss for a single noised pair as
\begin{equation*}
\mathcal{L}_\theta(x_0,x_t) := \sum_{k=1}^d -\log p_\theta\big(x_0^{(k)}, k \mid x_t\big).
\end{equation*}
The loss $\mathcal{L}_t(\theta)$ derived above is recovered as $\mathcal{L}_t(\theta) = \mathbb{E}_{x_0\sim q_0,\,x_t\sim q_t(\cdot\mid x_0)}\big[\mathcal{L}_\theta(x_0,x_t)\big]$.

\begin{proposition*}
If $p_\theta$ is $G$-invariant, i.e.\ $p_\theta(s,\,g\cdot k \mid g\cdot x_t) = p_\theta(s,k\mid x_t)$ for all $g\in G$, then $\mathcal{L}_\theta(x_0,x_t) = \sum_{k=1}^d -\log p_\theta(x_0^{(k)},k\mid x_t)$ is $G$-invariant as well:

\begin{equation*}
\mathcal{L}_\theta(g\cdot x_0,\, g\cdot x_t) = \mathcal{L}_\theta(x_0, x_t), \qquad \forall g\in G.
\end{equation*}
This holds for any noising discrete diffusion process, since the claim concerns only $p_\theta$ and the combinatorics of the sum, not $q$.
\end{proposition*}
\begin{proof}
We know $(g\cdot x_0)^{(k)} = x_0^{(g^{-1}\cdot k)}$, so
\begin{equation*}
\mathcal{L}_\theta(g\cdot x_0, g\cdot x_t) = \sum_{k=1}^d -\log p_\theta\big(x_0^{(g^{-1}\cdot k)},\, k \mid g\cdot x_t\big).
\end{equation*}
Writing $j := g^{-1}\cdot k$ (so $k=g\cdot j$) and setting $s=x_0^{(j)}$ in the equivariance hypothesis, $p_\theta\big(x_0^{(j)},\, g\cdot j \mid g\cdot x_t\big) = p_\theta\big(x_0^{(j)}, j \mid x_t\big)$, hence each term becomes
\begin{equation*}
-\log p_\theta\big(x_0^{(g^{-1}\cdot k)}, k \mid g\cdot x_t\big) = -\log p_\theta\big(x_0^{(j)}, j \mid x_t\big).
\end{equation*}
Since $k\mapsto j=g^{-1}\cdot k$ is a bijection of $\{1,\dots,d\}$, summing over $k$ reindexes the sum without changing its value:
\begin{equation*}
\mathcal{L}_\theta(g\cdot x_0, g\cdot x_t) = \sum_{k=1}^d -\log p_\theta\big(x_0^{(g^{-1}\cdot k)}, k\mid g\cdot x_t\big) = \sum_{j=1}^d -\log p_\theta\big(x_0^{(j)}, j\mid x_t\big) = \mathcal{L}_\theta(x_0,x_t).
\end{equation*}
\end{proof}

\subsection{Invariant training is equivalent to data augmentation}

Now that we have established the group action over tokens, 
we will first prove that using an equivariant network is equivalent to performing data augmentation on the un-symmetrized empirical data (Proposition~\ref{proposition_3}).  We will then use this result to derive the local-equivariant UDM optimum (Proposition~\ref{prop: prop5}). In Appendix~\ref{app:GIDD} we extend it to the more general GIDD framework.

\begin{proposition}[Invariant training is equivalent to data augmentation] \label{proposition_3}
Let $q_0(x) := \frac{1}{|D|}\sum_{x_0' \in D}\delta_{x_0'}(x)$ be the empirical dataset distribution and
$\bar q_{0, G}(x) := \frac{1}{|G|}\sum_{g\in G} q_0(g^{-1}\cdot x)$
its $G$-augmented version. If $\mathcal{L}_\theta$ satisfies the invariance condition, then
\begin{equation}
\mathbb{E}_{q_0(x_0),\ q_t(x_t \mid x_0)}\big[\mathcal{L}_\theta(x_0,x_t)\big] \;=\; \mathbb{E}_{ \bar q_{0, G}(x_0),\ q_t(x_t \mid x_0)}\big[\mathcal{L}_\theta(x_0,x_t)\big].
\label{eq:equal-losses}
\end{equation}
In words: training $p_\theta$ on the raw dataset $D$ is equivalent, in expectation, to training on the $G$-augmented dataset.
\end{proposition}

\begin{proof}
Fix $g\in G$. By assumption, $\mathcal{L}_\theta(g\cdot x_0, g\cdot x_t) = \mathcal{L}_\theta(x_0,x_t)$ for all $x_0,x_t$; in particular this holds for every $g\in G$, so
\begin{equation}
\mathcal{L}_\theta(x_0,x_t) = \frac{1}{|G|}\sum_{g\in G}\mathcal{L}_\theta(g\cdot x_0,\, g\cdot x_t).
\label{eq:trivial-avg}
\end{equation}
Using \eqref{eq:trivial-avg} and $q_0(x_0) = \frac{1}{|D|}\sum_{x_0'\in D}\delta_{x_0'}(x_0)$,
\begin{align*}
\mathbb{E}_{q_0(x_0,x_t)}\big[\mathcal{L}_\theta(x_0,x_t)\big]
&= \sum_{x_0,x_t} q_0(x_0)\, q_t(x_t\mid x_0)\; \frac{1}{|G|}\sum_{g\in G}\mathcal{L}_\theta(g\cdot x_0,\,g\cdot x_t) \\
&= \frac{1}{|G|}\sum_{g\in G}\ \sum_{x_0,x_t} q_0(x_0)\, q_t(x_t\mid x_0)\, \mathcal{L}_\theta(g\cdot x_0,\,g\cdot x_t).
\end{align*}
For fixed $g$, the map $(x_0,x_t)\mapsto(y_0,y_t):=(g\cdot x_0,\,g\cdot x_t)$ is a bijection of $\mathcal V^d\times\mathcal V^d$ with inverse $(y_0,y_t)\mapsto(g^{-1}\cdot y_0,\,g^{-1}\cdot y_t)$. Reindexing the inner sum by $(y_0,y_t)$,
\begin{equation*}
\sum_{x_0,x_t} q_0(x_0)\,q_t(x_t\mid x_0)\,\mathcal{L}_\theta(g\cdot x_0,\,g\cdot x_t)
= \sum_{y_0,y_t} q_0(g^{-1}\cdot y_0)\, q_t\big(g^{-1}\cdot y_t \mid g^{-1}\cdot y_0\big)\, \mathcal{L}_\theta(y_0,y_t).
\end{equation*}
Using $g^{-1}\in G$ and $q_t\big(g^{-1}\cdot y_t \mid g^{-1}\cdot y_0\big) = q_t(y_t\mid y_0)$, this becomes
\begin{equation*}
\sum_{y_0,y_t} q_0(g^{-1}\cdot y_0)\, q_t(y_t\mid y_0)\, \mathcal{L}_\theta(y_0,y_t).
\end{equation*}
Substituting back and exchanging the (finite) sum over $g$ with the sum over $(y_0,y_t)$,
\begin{align*}
\mathbb{E}_{q_0(x_0,x_t)}\big[\mathcal{L}_\theta(x_0,x_t)\big]
&= \sum_{y_0,y_t} \left(\frac{1}{|G|}\sum_{g\in G} q_0(g^{-1}\cdot y_0)\right) q_t(y_t\mid y_0)\, \mathcal{L}_\theta(y_0,y_t) \\
&= \sum_{y_0,y_t} \bar q_{0,G}(y_0)\, q_t(y_t\mid y_0)\, \mathcal{L}_\theta(y_0,y_t) \;=\; \mathbb{E}_{\bar q_{0,G}(x_0,x_t)}\big[\mathcal{L}_\theta(x_0,x_t)\big],
\end{align*}
which is \eqref{eq:equal-losses}.
\end{proof}

If we choose $G$ to be the \emph{translation group $\mathcal{T}$}, this result implies that training is agnostic to coordinates as we are virtually training on the augmented dataset $\mathcal{T}(D)$. In particular, when the model sees the dataset patches $\Ntk$, the augmented dataset of patches $\mathcal{N}(D)$ turns out to be the collection of all (weighted by frequency) seen patches along the dataset. As for any patch in some coordinate $k'$: $N_{t,k'}$ there exists a $g\in G$ s.t.\ $g\cdot N_{t,k'}=\Ntk$.

This leads to the Boltzmann kernel construction for UDM presented in the following Proposition.

\begin{proposition}[Translation equivariance] \label{prop: prop5}
For a local-equivariant Uniform Diffusion model (that is, following Eq.~\eqref{eq:udm-local} and Eq.~\eqref{eq:equivariance}) the loss optimum is given by the unanchored local posterior:

\begin{equation*}
p_\theta^{\star}\big(x_0^{(k)} \mid \Ntk\big) = q_{\mathrm{emp}}\big(x_0^{(k)} \mid \Nt\big)
\end{equation*}

in particular it has the following expression:

\begin{equation*}
\small
p_\theta^{\star}\big(x_0^{(k)}{=}c \mid \Nt\big)
=
\frac{
    \sum_{N_0 \in \Ndata}
    \ind{\mathrm{center}(N_0){=}c}\,
    e^{-\beta_t \dH(N_0,\Nt)}
}{
    \sum_{N_0' \in \Ndata}
    e^{-\beta_t \dH(N_0',\Nt)}
},
\quad
\beta_t =
\log\!\left(1+\frac{\alpha_t K}{1-\alpha_t}\right).
\end{equation*}
\end{proposition}
\begin{proof}
We apply Bayes' theorem and expand over the amplified dataset which we consider the training one, given training loss equivariance:
\begin{equation*}
q_{\mathrm{emp}}(x_0^{(k)}{=}c\mid\Nt) = \frac{\sum_{N_0\in\Ndata}\ind{\mathrm{center}(N_0)=c}\,q(\Nt\mid N_0)}{\sum_{N_0'\in\Ndata} q(\Nt\mid N_0')}.
\end{equation*}
where the forward corruption process is:

\begin{equation*}
\begin{aligned}
q(\Nt\mid N_0)
&= \Big(\alpha_t+\tfrac{1-\alpha_t}{K}\Big)^{m}
   \Big(\tfrac{1-\alpha_t}{K}\Big)^{w^2-m}
 = \Big(\tfrac{1-\alpha_t}{K}\Big)^{w^2}
   \Big(1+\tfrac{\alpha_t K}{1-\alpha_t}\Big)^{m} \\
&= \Big(\tfrac{1-\alpha_t}{K}\Big)^{w^2}
   e^{\beta_t w^2}\,
   e^{-\beta_t\,\dH(N_0,\Nt)} .
\end{aligned}
\end{equation*}

where $m:=w^2-\dH(N_0,\Nt)$ is the number of positions at which $N_0$ and $\Nt$ agree, and $\beta_t:=\log\!\big(1+\alpha_t K/(1-\alpha_t)\big)$. The prefactor $\big(\tfrac{1-\alpha_t}{K}\big)^{w^2}e^{\beta_t w^2}$ is the same for every $N_0\in\Ndata$, so it cancels in the Bayes ratio. Because the amplified dataset seamlessly considers any patch we are no longer anchored at position $k$, and therefore can claim: $p_\theta^{\star}\big(x_0^{(k)} \mid \Ntk\big) = q_{\mathrm{emp}}\big(x_0^{(k)} \mid \Nt\big)$.
\end{proof}

\section{Experimental results: uniform diffusion}
\label{app:exp}

\subsection{Setup}

Validating our analytical theory against a trained model requires comparing two conditional distributions over the $K$ classes: the model prediction $p_\theta(x_0^{(k)}\mid \Nt)$ and the analytical prediction $q_{\mathrm{theory}}(x_0^{(k)}\mid \Nt)$. For each query $q$, we compute three statistics:
\begin{equation}
KL_q
:= \sum_{c\in\mathcal{V}} p_\theta(c)
\log\frac{p_\theta(c)}{q_{\mathrm{theory}}(c)},
\qquad
TV_q
:= \tfrac12\sum_{c\in\mathcal{V}}
\big|p_\theta(c)-q_{\mathrm{theory}}(c)\big|,
\qquad
\mathrm{Top1}_q
:= \ind{\hat c_\theta=\hat c_q},
\label{eq:metrics}
\end{equation}
where
\[
\hat c_\theta := \arg\max_{c\in\mathcal{V}} p_\theta(c),
\qquad
\hat c_q := \arg\max_{c\in\mathcal{V}} q_{\mathrm{theory}}(c).
\]

These metrics capture complementary notions of agreement. KL measures distributional discrepancy and is directly aligned with the cross-entropy training objective, but is unbounded above and sensitive to discrepancies in low-probability regions. TV lies in $[0,1]$ and directly quantifies the amount of probability mass that must be redistributed to make the two distributions coincide. Top-1 agreement measures whether model and theory select the same token under greedy decoding.

We average each metric over $Q=500$ held-out queries sampled across $(C,H,W)$, independently at each of five noise levels,
\[
\alpha_t \in \{0.9,0.7,0.5,0.3,0.1\}.
\]
Empirically, $Q=500$ yields stable estimates: even in the worst case, the resulting averages remain within a $10\%$ relative error of the reference mean, which is sufficient for our purposes.

\subsection{Model training}

We train UDM models whose receptive field is a $w\times w$ square field with circular padding, for a small $w$ value. By the use of circular  padding, we will not evaluate the impact of boundary conditions on generative models and equivariance properties. We estimate that on big dimensions equivariance is approximately true even when boundaries are present. We refer the reader to boundary treatment for analytical equivariant theories in continuous diffusion models \cite{kamb}.

\textbf{Training objective.} Models are trained with the standard cross-entropy loss,
\begin{equation*}
\mathcal{L}(\theta) = \mathbb{E}_{t \sim U(0,1)}\ \mathbb{E}_{x_t \sim q(\cdot \mid x_0)} \left[ -\frac{1}{|D|} \sum_{x_0 \in D} \frac{1}{H W} \sum_{k=1}^{HW} \log p_\theta\big(x_0^{(k)} \mid \Ntk\big) \right].
\end{equation*}

The average runs over every position since at any noise level every position carries partial information about its own clean token. Optimization uses Adam~\cite{adam} with learning rate $10^{-3}$, batch size $64$, and $3000$ steps, with $t$ drawn uniformly per batch and $\alpha_t = 1 - t$.

\textbf{Datasets.} We train on $N{=}5000$ images per configuration and evaluate on $500$ images held out from the corresponding test split. The datasets we use are:
\begin{itemize}
\item \textbf{MNIST}~\cite{mnist} ($28\times28$, $K{=}2$): handwritten digits binarized at intensity $0.5$.
\item \textbf{Fashion-MNIST}~\cite{fashionmnist} ($28\times28$, $K{=}2$): clothing silhouettes, binarized identically. A drop-in replacement for MNIST at the same resolution and vocabulary.
\item \textbf{CIFAR-10}~\cite{Krizhevsky} ($32\times32$, $K\in\{2,4,8,\dots\}$): natural photographs converted to greyscale with standard luminance weights.
\end{itemize}

\begin{figure}[h]
\centering
\includegraphics[width=\textwidth]{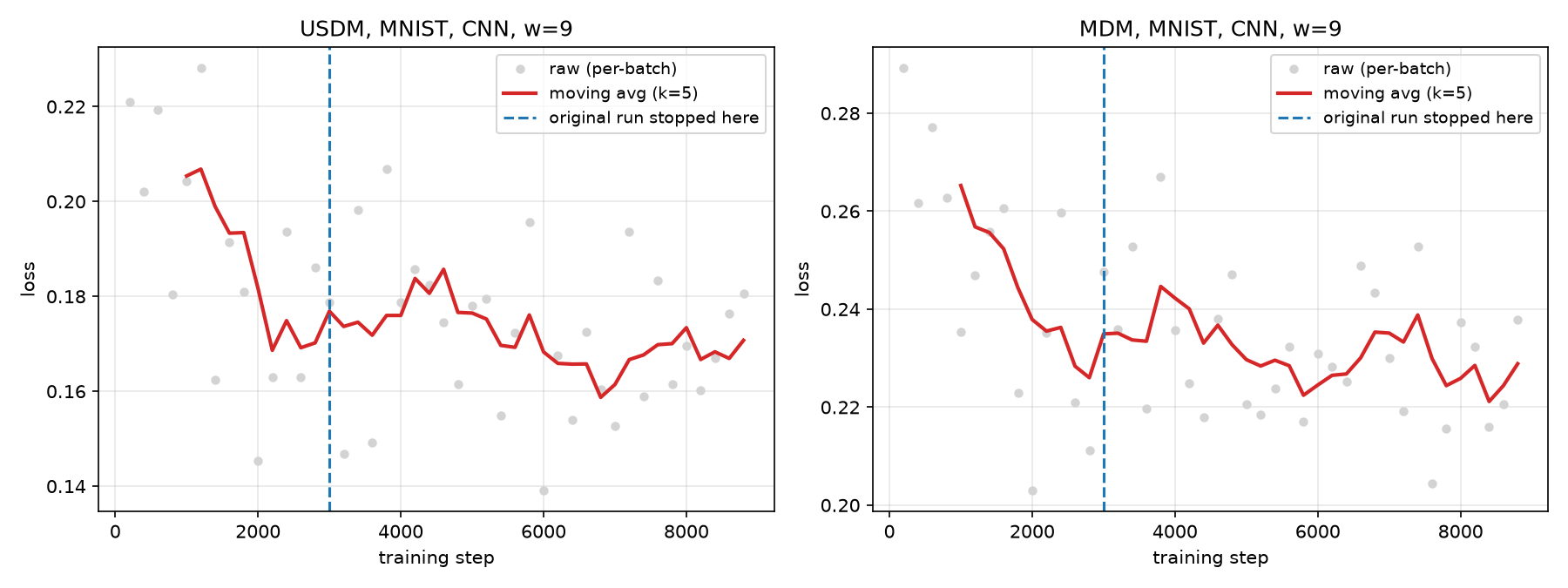}
\caption{Per-step training loss (gray) and its moving average (red) for the largest window in the sweep on MNIST, trained three times past the budget used for every reported result. The dashed line marks where the reported runs stop. Left: UDM. Right: MDM.}
\label{fig:convergence}
\end{figure}

We sweep $w \in \{3,5,7, 9\}$ and, at each, compute Eq.~\eqref{eq:metrics} against the closed form derived from that same training set. In order to justify architecture independence of our claim we measure against three distinct architectures: CNN, GNN and NATTEN, obtaining similar results for all. We train using $3000$ steps, which have proven to be enough for convergence in the worst case we test; see Figure~\ref{fig:convergence}.

\subsection{Main results}

\begin{table}[ht!]
\centering
\resizebox{\textwidth}{!}{%
\begin{tabular}{lccccccccc}
\toprule
& \multicolumn{3}{c}{$w{=}3$} & \multicolumn{3}{c}{$w{=}5$} & \multicolumn{3}{c}{$w{=}7$} \\
\cmidrule(lr){2-4}\cmidrule(lr){5-7}\cmidrule(lr){8-10}
Dataset & KL & TV & Top-1 & KL & TV & Top-1 & KL & TV & Top-1 \\
\midrule
MNIST ($K{=}2$)         & 0.0009 & 0.0065 & 99.56\% & 0.0057 & 0.0137 & 99.12\% & 0.0104 & 0.0172 & 98.68\% \\
Fashion-MNIST ($K{=}2$) & 0.0022 & 0.0152 & 98.12\% & 0.0066 & 0.0206 & 98.28\% & 0.0102 & 0.0196 & 98.16\% \\
CIFAR-10 ($K{=}2$)      & 0.0013 & 0.0113 & 98.40\% & 0.0038 & 0.0170 & 97.52\% & 0.0144 & 0.0241 & 97.48\% \\
CIFAR-10 ($K{=}4$)      & 0.0060 & 0.0248 & 97.20\% & 0.0269 & 0.0452 & 95.04\% & 0.0991 & 0.0740 & 91.80\% \\
CIFAR-10 ($K{=}8$)      & 0.0244 & 0.0514 & 92.24\% & 0.1382 & 0.1014 & 89.12\% & 1.6989 & 0.1758 & 82.48\% \\
\bottomrule
\end{tabular}%
}

\vspace{8pt}
\caption{Mean KL, TV and top-1 agreement between the trained model (CNN) and the analytical expression Eq.~\eqref{eq:optimum} for $N{=}5000$ training samples. Reproduced from the main text.}
\label{tab:app-usdm-main}
\end{table}

It is worth commenting that the three metrics degrade very unevenly: KL is the most sensitive of them and is easily altered by small mismatches, which a trained softmax can never fully avoid since it never assigns exactly zero mass; TV and top-1 agreement move more gently across the same rows. A large KL at the harder end of the grid should therefore be read as a sign that the comparison is stressed, not that the model's predictions have become unrelated to the theory, since in those same cells the model and the closed form still choose the same token in the vast majority of positions.

Even with the decay depending most notably on vocabulary size $K$, an 80\% top-1 agreement is still worth noting. The degradation indicates that the current theory is disregarding a factor that simplifies itself at low $K$ and $w$ and that becomes relevant for the other regions, being a foundational explanation lacking pieces to describe the whole of the mechanism. 

\subsection{Cross-Architecture results} \label{app: cross arch results}

We find that our claims hold for various architectures similarly. We train against CNN, GNN and NATTEN. 

\begin{itemize}
    \item \textbf{CNN}~\cite{cohen2016groupequivariantconvolutionalnetworks} applies $w\times w$ convolution with circular padding.
    \item \textbf{GNN}~\cite{gilmer2017neuralmessagepassingquantum} treats the window as a neighbourhood in a periodic lattice graph: a message function is applied to every centre-neighbour pair, the resulting messages are averaged, and a shared update function produces the output implying that the same weights act on every neighbour regardless of where it sits. 
    \item \textbf{NATTEN}~\cite{hassani2023neighborhoodattentiontransformer} restricts self-attention to the $w\times w$ neighbourhood, with circular padding applied before attention and cropped afterwards, so that attention weights are computed from content similarity between the centre and each neighbour.
\end{itemize}

\begin{table}[h]
\centering
\begin{tabular}{llcccc}
\toprule
& & \multicolumn{4}{c}{$w$} \\
\cmidrule(lr){3-6}
Dataset & Architecture & 3 & 5 & 7 & 9 \\
\midrule
MNIST
& CNN    & 0.0009 & 0.0057 & 0.0104 & 0.0205 \\
& NATTEN & 0.0021 & 0.0300 & 0.0927 & 0.1706 \\
& GNN    & 0.0020 & 0.0303 & 0.0942 & 0.1744 \\
\addlinespace
Fashion-MNIST
& CNN    & 0.0022 & 0.0066 & 0.0102 & 0.0246 \\
& NATTEN & 0.0060 & 0.0229 & 0.0644 & 0.1297 \\
& GNN    & 0.0054 & 0.0228 & 0.0620 & 0.1299 \\
\addlinespace
CIFAR-10 ($K{=}2$)
& CNN    & 0.0013 & 0.0038 & 0.0144 & 0.0493 \\
& NATTEN & 0.0044 & 0.0252 & 0.0618 & 0.1486 \\
& GNN    & 0.0039 & 0.0249 & 0.0617 & 0.1474 \\
\bottomrule
\end{tabular}

\vspace{8pt}
\caption{Mean KL between the trained model and Eq.~\eqref{eq:optimum} for the three local, translation-equivariant backbones, $N{=}5000$, identical protocol in every row.}
\label{tab:app-cross-arch}
\end{table}

We see that results are very similar across architectures. The best performing one is CNN and the worst performing GNN, although the difference is almost negligible. For bigger windows the difference grows, especially comparing CNN against NATTEN and GNN.

\subsection{Ablation study against $K$}

\begin{table}[h]
\centering
\begin{tabular}{lccccc}
\toprule
$K$ & 2 & 4 & 8 & 16 & 32 \\
\midrule
KL     & 0.0013 & 0.0060 & 0.0244 & 0.0598 & 0.1521 \\
TV     & 0.0113 & 0.0248 & 0.0514 & 0.0807 & 0.1412 \\
Top-1  & 98.40\% & 97.20\% & 92.24\% & 90.16\% & 81.88\% \\
\bottomrule
\end{tabular}

\vspace{8pt}
\caption{Mean KL, TV and top-1 agreement at fixed $w{=}3$ on quantized greyscale CIFAR-10, CNN, $N{=}5000$, as the vocabulary size $K$ grows.}
\label{tab:app-k-sweep}
\end{table}

We fix the best performing $w$ size and sweep against $K$. For a low $w=3$, which is the optimal in Kamb \& Ganguli's work for an analogous continuous diffusion model theory, the match of theory vs model extends to higher values of $K$.  The trend is coherent with our data signal-to-noise ratio in $K^{w^2}$ combinatorial space heuristic.

\subsection{Match vs dataset size}

In order to test if dataset size does generate more signal for the theory in $K^{w^2}$ space we compare the match for different sizes. The results are shown in Table~\ref{tab:app-n-scaling}.

\begin{table}[h]
\centering
\begin{tabular}{lcccccc}
\toprule
& \multicolumn{3}{c}{$N{=}5000$} & \multicolumn{3}{c}{$N{=}30000$} \\
\cmidrule(lr){2-4}\cmidrule(lr){5-7}
$w$ & KL & TV & Top-1 & KL & TV & Top-1 \\
\midrule
3 & 0.0060 & 0.0248 & 97.20\% & 0.0055 & 0.0233 & 97.04\% \\
5 & 0.0269 & 0.0452 & 95.04\% & 0.0254 & 0.0442 & 94.68\% \\
7 & 0.0991 & 0.0740 & 91.80\% & 0.0767 & 0.0618 & 93.84\% \\
9 & 1.4457 & 0.1233 & 89.16\% & 1.0370 & 0.1028 & 90.52\% \\
\bottomrule
\end{tabular}

\vspace{8pt}
\caption{Mean KL, TV and top-1 agreement on quantized CIFAR-10 ($K{=}4$), CNN, at $N{=}5000$ against $N{=}30000$ training images.}
\label{tab:app-n-scaling}
\end{table}

More data does help: at small windows where the match was already close this extra signal is not useful and leaves the match approximately the same. At the large windows, where the discrepancy is largest, the same increase produces a reduction of nearly a third in KL, which denotes the whole distribution match. Top-1 and TV, which were already high, do not move substantially. This is of interest to us because TV and Top-1 describe the broad behaviour of the prediction, and KL its exact match. 

Although the improvement is measurable, it is modest, and the numerical evidence points to a missing ingredient in the theory: closing the gap through data alone would require the dataset to grow exponentially with $w^2$, which is not realistic.

\section{Experimental results: masked diffusion}
\label{app:exp-mdm}

\subsection{What is being tested}

After analyzing the local equivariant optimum for the Masked Diffusion case we obtain the following expression (proof in Appendix~\ref{app:GIDD}):

\begin{equation}
p_\theta^{\star}\big(x_0^{(k)} \mid \Nt\big) = q_{\mathrm{emp}}\big(x_0^{(k)} \mid \Nt^{\mathrm{vis}}\big)
= \frac{\displaystyle\sum_{N_0 \in \Ndata} \ind{N_0^{\mathrm{vis}}{=}\Nt^{\mathrm{vis}}}\,\ind{\mathrm{center}(N_0){=}x_0^{(k)}}}
       {\displaystyle\sum_{N_0' \in \Ndata} \ind{N_0^{\prime\,\mathrm{vis}}{=}\Nt^{\mathrm{vis}}}},
\label{eq:mdm-optimum}
\end{equation}

This expression considers dataset patches that match the current revealed tokens. This expression is clearly undefined for revealed tokens nowhere seen in any patch in the dataset.

Consider a dataset consisting of only two samples: one entirely black and one entirely white. Under ancestral sampling~\cite{shi2025simplifiedgeneralizedmaskeddiffusion}, locality allows spatially distant tokens to evolve independently, so an intermediate sample may contain both black and white regions. Consequently, for some token $k$, its visible neighborhood $N_t^{\mathrm{vis}}$ may contain both black and white tokens. Such a neighborhood never occurs in the training set, whose local patches are either entirely black or entirely white. An exact lookup rule therefore assigns zero weight to both possibilities: the all-white patch is incompatible with the observed black tokens, while the all-black patch is incompatible with the observed white tokens. Hence the resulting conditional distribution cannot be normalized and is undefined for patches outside the support of the training data.

A more reasonable behavior would be to interpolate beyond exact matches and quantify which training pattern is more compatible with the observed neighborhood. In this example, one would naturally expect the prediction to favor the majority state: if most visible tokens are black, predict black, and analogously for white.

Given that the \emph{exact posterior} is the limit of the UDM Boltzmann kernel when $\beta_t \to \infty$, a natural way to incorporate the relaxation associated with finite-step training is to consider that the actual optimum has the shape: 

\begin{equation}
q_{\hat\beta}\big(x_0^{(k)}{=}c \mid \Nt\big)
= \frac{\displaystyle\sum_{N_0 \in \Ndata} \ind{\mathrm{center}(N_0){=}c}\; e^{-\hat\beta\, \dH^{\mathrm{vis}}(N_0, \Nt)}}
       {\displaystyle\sum_{N_0' \in \Ndata} e^{-\hat\beta\, \dH^{\mathrm{vis}}(N_0', \Nt)}}.
\label{eq:app-mdm-beta}
\end{equation}

We emphasize that this is a heuristic rather than a formal statement; a rigorous mathematical justification is left for future work.

The relaxed expression is defined for every input and recovers the exact rule as $\hat\beta \to \infty$. The essential difference from the uniform case is that $\hat\beta$ has no closed form in terms of $\alpha_t$ and $K$: the masked corruption process itself fixes the inverse temperature at infinity. The finite value observed in a trained model is therefore a property of its parameterization and training, not of the diffusion dynamics.

Because $\hat\beta$ is fitted rather than derived, a close fit is weaker evidence than it may appear: a one-parameter family could, in principle, fit any converged model, given the space of functions depending on $\hat\beta$ we consider is general enough. In Section~\ref{app:mdm-check} we test whether this is the case.

\subsection{Method}

We train using the cross-entropy loss on masked positions only as the dynamic imposes \cite{shi2025simplifiedgeneralizedmaskeddiffusion, sahoo2024simpleeffectivemaskeddiffusion},
\begin{equation*}
\mathcal{L}_{\mathrm{MDM}}(\theta) = \mathbb{E}_{t \sim U(0,1)}\ \mathbb{E}_{z_t \sim q(\cdot \mid x_0)} \left[ \frac{1}{t}\cdot\frac{1}{|D|} \sum_{x_0 \in D} \frac{1}{HW} \sum_{k \,:\, z_t^{(k)} = \texttt{MASK}} -\log p_\theta\big(x_0^{(k)} \mid \Ntk\big) \right],
\end{equation*}
with the weight $w(t)=1/t$ that the continuous-time NELBO of the absorbing-state kernel requires under the linear schedule. Optimization is unchanged: Adam~\cite{adam} at $10^{-3}$, batch size $64$, $3000$ steps (already justified in Appendix~\ref{app:exp}).

We evaluate the following matches:

\begin{enumerate}
\item \textbf{The exact posterior}, Eq.~\eqref{eq:mdm-optimum}, the ideal optimum of the masked objective. It is evaluated only where some training patch reproduces the visible context exactly and it is matched against a converged model.
\item \textbf{The relaxed kernel}, Eq.~\eqref{eq:app-mdm-beta} at the $\hat\beta$ that best fits it. We evaluate $\hat\beta$ at fourteen candidate values spanning $[0,32]$, score each by the mean $KL(p_\theta \,\|\, q_{\hat\beta})$ over the same queries and keep the best one.
\end{enumerate}

\subsection{Results}
\begin{table}[h]
\centering
\begin{tabular}{llccc}
\toprule
Dataset & $w$ & KL (exact) & $\hat\beta$ & KL at $\hat\beta$ \\
\midrule
MNIST         & 3 & 0.0011 & 9.6 & 0.0015 \\
              & 5 & 0.0440 & 6.6 & 0.0102 \\
              & 7 & 0.3410 & 2.5 & 0.0141 \\
\addlinespace
Fashion-MNIST & 3 & 0.0034 & 6.4 & 0.0033 \\
              & 5 & 0.1814 & 6.4 & 0.0114 \\
              & 7 & 0.4067 & 2.9 & 0.0278 \\
\addlinespace
CIFAR-10 ($K{=}2$) & 3 & 0.0018 & 6.0 & 0.0017 \\
                   & 5 & 0.1526 & 4.6 & 0.0091 \\
                   & 7 & 0.6041 & 2.4 & 0.0246 \\
\addlinespace
CIFAR-10 ($K{=}4$) & 3 & 0.0301 & 5.2 & 0.0139 \\
                   & 5 & 1.4868 & 2.8 & 0.0496 \\
                   & 7 & 2.1015 & 1.8 & 0.0867 \\
\addlinespace
CIFAR-10 ($K{=}8$) & 3 & 0.7101 & 4.4 & 0.0405 \\
& 5 & 4.2007 & 2.5 & 0.1056 \\
& 7 & 3.3673 & 1.6 & 0.1861 \\
\bottomrule
\end{tabular}

\vspace{8pt}
\caption{Masked diffusion, CNN backbone, $N{=}5000$. KL (exact) is the mean divergence from the trained model to Eq.~\eqref{eq:mdm-local-equiv-opt}, evaluated on the queries whose visible context is matched exactly by a training patch. $\hat\beta$ is the grid-search fit of Eq.~\eqref{eq:app-mdm-beta} and the last column the mean KL it achieves on the same queries.}
\label{tab:app-mdm-main}
\end{table}

The match against the exact kernel deteriorates quickly as $w$ and $K$ grow, whereas the relaxed kernel remains close to the trained model on every dataset. For small windows and vocabularies the two achieve similar values. Note, however, that this comparison is restricted to queries whose visible context appears in the training set, the only ones on which the exact kernel is defined. The relaxed kernel is also defined on unseen contexts, so it describes the model's behaviour more generally.

Although it outperforms the exact posterior, the relaxed kernel still degrades as $K$ and $w$ increase. We attribute this to the limited coverage of the combinatorial space $K^{w^2}$ by the dataset, as discussed in Appendix~\ref{app:exp}.

\subsection{Is the relaxed kernel picking up real structure?}
\label{app:mdm-check}

A fitted one-parameter family that lands close to a trained model is not by itself evidence that the model learned anything about Hamming distance to training patches. It could happen that Eq.~\eqref{eq:app-mdm-beta} is a sufficiently flexible family of functions that some value of $\hat\beta$ encompasses almost any distribution a confident classifier might produce. To show that this is not the case we test the fit against an explicit null and against an untrained control for the dataset MNIST using $w{=}5$ at the middle noise level.

\textbf{Tests.}

\begin{enumerate}
\item \textbf{Label-permutation null.} Every training patch is kept as it is originally, while the assignment of centre classes to patches is randomized. The objective is to test the Hamming distance proportionality. The visible contexts and the class frequencies are therefore unchanged and only the association between a patch's neighbourhood and its centre is destroyed. If the model's confidence bears no real relationship to which patches lie close to a query, the fit should survive this. We repeat the permutation a thousand times, refitting $\hat\beta$ each time.
\item \textbf{Untrained control.} Three randomly initialized networks of the same architecture are evaluated against the same queries and patches. This asks the complementary question: is the local equivariant architecture alone enough to produce a good fit without any training?
\end{enumerate}

\textbf{Results.} Table~\ref{tab:app-mdm-null} reports the fit for the trained model and for the two controls.

\begin{table}[ht!]
\centering
\begin{tabular}{lcc}
\toprule
& Mean KL at $\hat\beta$ & Trials \\
\midrule
Trained model            & 0.0158 & 1 \\
Permuted null            & $0.3113 \pm 0.0012$ & 1000 \\
Untrained networks       & $0.4383 \pm 0.0531$ & 3 \\
\bottomrule
\end{tabular}

\vspace{8pt}
\caption{Significance checks at MNIST, $w{=}5$, $\alpha_t{=}0.5$. Each row reports the mean KL achieved by the best-fitting $\hat\beta$, with the standard deviation across trials where more than one was run.}
\label{tab:app-mdm-null}
\end{table}

The results show us that both permuted and untrained networks have a drastically higher KL than converged trained models, providing supporting evidence that this match is indeed a consequence of the masked model's creative and interpolating behaviour and not a consequence of the possible universality of our proposed family of distributions. Both the permuted and the untrained networks' KL to the model lie several standard deviations away from the actual trained model, whose match is exceptionally close.

\section{Theoretical proofs for the GIDD framework}
\label{app:GIDD}

In this section, we present the theoretical proofs for the Generalized Interpolating Discrete Diffusion (GIDD) framework~\cite{vonrütte2025generalizedinterpolatingdiscretediffusion}, which allows for interpolation between Uniform and Masked. These results are valuable for future explorations tackling learning and creative mechanics as the masked level is shifted. We take advantage of the results we have on the GIDD framework to present the masked diffusion proofs as corollaries, as they are immediate, once GIDD results are derived.

\begin{remark*}
One would think that this framework would represent the UDM dynamic as well; however, we are interested in the Cross Entropy loss training of UDM, which is not equivalent to the GIDD loss. For the MDM case it does accurately generalize, however.
\end{remark*}

Working on the GIDD framework we are going to use the following notation:

\textbf{Notation.} A model of the distribution of a noisy state $q_t(x_t | x_0)$ using $x_\theta(x_t)$ as a plug-in estimator of $x_0$ will be denoted by $q_\theta (\cdot\;; x_t)\equiv q_t(\cdot |x_\theta(x_t))$, where we have included $x_t$ as an argument to emphasize the model behaviour as a function from a noisy sample $x_t$, which will become important in later proofs. Note that we write $x_\theta$ to ease interpretation, but it in fact coincides with the probability modeled posterior $x_\theta(x_t) \equiv p_\theta(\cdot|x_t)$. We will call $q_\theta (\cdot\;; x_t)$ the \emph{plug-in noisy marginal}.

The GIDD ELBO for a one-dimensional token is the following~\cite{vonrütte2025generalizedinterpolatingdiscretediffusion}.

\begin{proposition}[GIDD ELBO]
\label{prop:gidd-elbo}
Let $\alpha_t$ and $\pi_t$ be a mixing schedule with marginal forward distribution $q_t(x_t\mid x) = \mathrm{Cat}(x_t;\, \alpha_t x + (1-\alpha_t) \pi_t)$. Let $q_\theta (x_0; x_t)$ denote the plug-in noisy marginal.
Define the weighting function
\begin{equation*}
w_t(x_t, x) = \frac{1}{q_t(x_t\mid x)}\, x_t^\top\!\left((1-\alpha_t) \pi_t' - \frac{\alpha_t'}{\alpha_t}\pi_t\right).
\end{equation*}
Then the continuous-time negative ELBO of the corresponding diffusion model is
\begin{align*}
\mathbb{E}_{x_0}\big[-\log p(x_0)\big] \;\leq\; \mathcal{L}(\theta) + C ={}& \mathbb{E}_{t, x_0,\,x_t}\Big[w_t(x_t,x)\Big(D_{\mathrm{KL}}\big(q_t(\cdot\mid x_0)\,\|\,q_\theta (\cdot\;; x_t)\big) \\
& + D_{\mathrm{IS}}\big(q_t(x_t\mid x_0)\,\|\,q_\theta (x_t; x_t)\big)\Big)\Big] + C,
\end{align*}
where $x_0\sim\mathcal U(D)$, $t\sim\mathcal U(0,1)$, $x_t \sim q_t(\cdot\mid x_0)$, $D_{\mathrm{IS}}(p\|q) = p/q - \log(p/q) - 1$ is the pointwise Itakura--Saito divergence, and
\begin{equation*}
C = \mathbb{E}_{q_0(z_0\mid x)}\big[\log p(x\mid z_0)\big] - D_{\mathrm{KL}}\big(q_1(z_1\mid x)\,\|\,p_1(x_1)\big)
\end{equation*}
is a schedule-dependent constant, independent of $\theta$.
\end{proposition}

\subsection{Multidimensional loss reformulation}

Now for the multi-dimensional case, we sum the loss over all coordinates $k\in \{1, \dots, d\}$ as the whole loss is based on forward marginals which factorise over dimensions. Letting our $k$-th token $x_t^{(k)}$ plug-in noisy marginal $q_\theta^{(k)}(\cdot\;; x_t)$ to be dependent however on the whole multidimensional input $x_t$ gives the following training objective:

\begin{equation*}
\mathcal{L}(\theta) = \mathbb{E}_{t \sim U(0,1)} \left[ \mathcal{L}_t(\theta) \right],
\end{equation*}
\begin{align}
\mathcal{L}_t(\theta) =\mathbb{E}_{q(x_0, x_t)} \bigg[ \sum_{k=1}^{d} w_t\big(x_t^{(k)}, x_0^{(k)}\big)\Big(&D_{\mathrm{KL}}\Big(q_t\big(\cdot \mid x_0^{(k)}\big)\,\Big\|\,q^{(k)}_\theta (\cdot\;; x_t)\Big) \nonumber\\
+\; &D_{\mathrm{IS}}\Big(q_t\big(x_t^{(k)} \mid x_0^{(k)}\big)\,\Big\|\,q^{(k)}_\theta (x_t^{(k)}; x_t)\Big)\Big) \bigg],
\label{eq:per-token-loss}
\end{align}

where the weighting is given by:

\begin{equation*}
w_t\big(x_t^{(k)}, x_0^{(k)}\big) = \frac{1}{q_t\big(x_t^{(k)}\mid x_0^{(k)}\big)}\, \big(x_t^{(k)}\big)^\top\!\left((1-\alpha_t) \pi_t' - \frac{\alpha_t'}{\alpha_t}\pi_t\right),
\end{equation*}
and where, for each position $k$, the model's plug-in noisy-marginal is

\begin{equation*}
q_\theta^{(k)}(\cdot\;; x_t) \;:=\; \alpha_t\, p^{(k)}_\theta\big(\cdot \mid x_t\big) + (1-\alpha_t)\, \pi_t,
\end{equation*}

i.e.\ exactly the same linear mixture that defines the true marginal $q_t(\cdot\mid x_0^{(k)}) = \alpha_t x_0^{(k)} + (1-\alpha_t) \pi_t$, but with the model's predicted distribution $p^{(k)}_\theta(\cdot\mid x_t)$ standing in for the (at inference time unknown) one-hot $x_0^{(k)}$. $q_t(\cdot \mid x_\theta)$ is a probability vector over the vocabulary, just like $p_\theta$ itself.

Noting that each term in the sum depends on $x_0$ only through $x_0^{(k)}$ we can take sum of expectations and marginalise over the rest of the tokens obtaining:
\begin{align*}
\mathcal{L}_t(\theta) =\sum_{k=1}^{d}\mathbb{E}_{q(x_0^{(k)}, x_t)} \bigg[  w_t\big(x_t^{(k)}, x_0^{(k)}\big)\Big(&D_{\mathrm{KL}}\Big(q_t\big(\cdot \mid x_0^{(k)}\big)\,\Big\|\,q^{(k)}_\theta (\cdot \;; x_t)\Big) \\
+\; &D_{\mathrm{IS}}\Big(q_t\big(x_t^{(k)} \mid x_0^{(k)}\big)\,\Big\|\,q^{(k)}_\theta (x_t^{(k)}; x_t)\Big)\Big) \bigg],
\end{align*}

\subsection{Bregman divergence properties}

The sum of $D_{KL}$ and $D_{IS}$ in the GIDD loss can be treated with a more general mathematical object: the Bregman loss~\cite{bregman}, as is also mentioned in the GIDD paper~\cite{vonrütte2025generalizedinterpolatingdiscretediffusion}.  

\begin{lemma}[$D_{\mathrm{KL}}$ and $D_{\mathrm{IS}}$ are Bregman divergences]
\label{lem:kl-is-bregman}
Let $D_\phi(a\|b) := \phi(a)-\phi(b)-\langle\nabla\phi(b),a-b\rangle$ denote the Bregman divergence of a differentiable, strictly convex potential $\phi$. Both $D_{KL}$ and $D_{IS}$ are Bregman divergences for the following potentials:
\begin{enumerate}
\item[(a)] $D_{\mathrm{KL}}(a\|b) = \sum_i a_i\log(a_i/b_i)$ is $D_{\phi_1}$ for $\phi_1(v):=\sum_i v_i\log v_i$.
\item[(b)] For fixed coordinate $z \in \mathcal{V}$, the pointwise $D_{\mathrm{IS}}(a_z\|b_z) = a_z/b_z-\log(a_z/b_z)-1$ is $D_{\phi_2}$ for $\phi_2(v):=-\log v_z$.
\end{enumerate}
\end{lemma}
\begin{proof}
(a) $\phi_1$ is strictly convex ($\partial^2\phi_1/\partial v_i^2=1/v_i>0$), with $\nabla\phi_1(v)_i=\log v_i+1$. For probability vectors $a,b$ ($\sum_i a_i=\sum_i b_i=1$),
\begin{equation*}
\langle\nabla\phi_1(b),a-b\rangle = \sum_i(\log b_i+1)(a_i-b_i) = \sum_i a_i\log b_i - \sum_i b_i\log b_i,
\end{equation*}
so $D_{\phi_1}(a\|b) = \sum_i a_i\log a_i - \sum_i a_i\log b_i = \sum_i a_i\log(a_i/b_i) = D_{\mathrm{KL}}(a\|b)$.

(b) $\phi_2$ is convex ($\partial^2\phi_2/\partial v_z^2=1/v_z^2>0$, zero elsewhere), with $\nabla\phi_2(v)=-\frac{1}{v_z}e_z$. Then $\langle\nabla\phi_2(b),a-b\rangle=-\frac{1}{b_z}(a_z-b_z)$, so
\begin{equation*}
D_{\phi_2}(a\|b) = -\log a_z+\log b_z+\frac{1}{b_z}(a_z-b_z) = \frac{a_z}{b_z}-\log\frac{a_z}{b_z}-1 = D_{\mathrm{IS}}(a_z\|b_z).
\end{equation*}
\end{proof}

Since Bregman divergences add when their generators add, $D_{\mathrm{KL}}(\cdot\|c)+D_{\mathrm{IS}}(\cdot_z\|c_z) = D_{\Phi,z}(\cdot\|c)$ for $\Phi:=\phi_1+\phi_2$, strictly convex on $\{v\in\Delta:v_z>0\}$ where $z = x_t^{(k)} \in \mathcal{V}$ (both $\phi_1$ and $\phi_2$ are convex so their sum is convex as well).

Going back to our loss it simplifies to:

\begin{equation}
\mathcal{L}_t(\theta) = \sum_{k=1}^{d} \mathbb{E}_{q\big(x_0^{(k)}, x_t\big)} \left[ w_t\big(x_t^{(k)}, x_0^{(k)}\big)\, D_{\Phi, x_t^{(k)}}\Big(q_t\big(\cdot \mid x_0^{(k)}\big) \,\Big\|\, q_\theta^{(k)}(\cdot\;; x_t)\Big) \right].
\label{eq:per-token-loss-bregman}
\end{equation}

\begin{lemma}[Weighted Bregman mean]
\label{lem:bregman-mean}
Let $\phi$ be a differentiable, strictly convex function on a convex domain, $X$ a random variable, and $w(X)\geq 0$ a weight with $\mathbb{E}[w(X)]>0$. Then
\begin{equation*}
\mathbb{E}\big[w(X)\,D_\phi(X\|c)\big] = \underbrace{\mathbb{E}[w(X)\phi(X)] - \mathbb{E}[w(X)]\,\phi(\bar X_w)}_{\text{independent of }c} \;+\; \mathbb{E}[w(X)]\cdot D_\phi(\bar X_w \| c),
\end{equation*}
where $\bar X_w := \mathbb{E}[w(X)X]/\mathbb{E}[w(X)]$. Consequently $c\mapsto \mathbb{E}[w(X)D_\phi(X\|c)]$ is uniquely minimized at $c=\bar X_w$.
\end{lemma}
\begin{proof}
Write $W:=\mathbb{E}[w(X)]$. Expanding and taking the expectation,
\begin{align*}
\mathbb{E}[w(X)D_\phi(X\|c)] &= \mathbb{E}[w(X)\phi(X)] - W\phi(c) - \big\langle \nabla\phi(c),\, \mathbb{E}[w(X)X] - Wc \big\rangle \\
&= \mathbb{E}[w(X)\phi(X)] - W\phi(c) - W\big\langle\nabla\phi(c),\,\bar X_w - c\big\rangle.
\end{align*}
Adding and subtracting $W\phi(\bar X_w)$ and recognizing $D_\phi(\bar X_w\|c)$ gives the stated identity. Since $D_\phi\geq 0$ with equality iff its arguments coincide, and $W>0$, the second term is minimized, at value $0$, uniquely at $c=\bar X_w$.
\end{proof}

\subsection{Locality in the GIDD framework}

\begin{proposition}[Locality, GIDD loss]
\label{prop:locality-gidd}
Under the restriction $p_\theta(x_0^{(k)}\mid x_t)\equiv p_\theta(x_0^{(k)}\mid N_{t,k})$, $\mathcal{L}_t(\theta)$ of Eq.~\eqref{eq:per-token-loss} is minimized, for every $N_{t,k}$, at
\begin{equation}
p_\theta^\star\big(x_0^{(k)} \mid N_{t,k}\big) \;=\; \frac{w_t\big(x_t^{(k)},\,x_0^{(k)}\,\big)\; q\big(x_0^{(k)} \mid N_{t,k}\big)}{\displaystyle\sum_{v\in\mathcal V} w_t\big(x_t^{(k)}, v\big)\; q\big(v \mid N_{t,k}\big)}.
\label{eq:locality-gidd-target}
\end{equation}
\end{proposition}
\begin{proof}
Because our model only depends on a neighborhood $\Ntk$, we can marginalise over non-neighbours, obtaining:

\begin{align*}
\mathcal{L}_t(\theta) &= \sum_{k=1}^{d} \mathbb{E}_{q\big(x_0^{(k)}, x_t\big)} \left[ w_t\big(x_t^{(k)}, x_0^{(k)}\big)\, D_{\Phi, x_t^{(k)}}\Big(q_t\big(\cdot \mid x_0^{(k)}\big) \,\Big\|\, q_\theta^{(k)}(\cdot\;; \Ntk)\Big) \right] \\
&= \sum_{k=1}^{d} \mathbb{E}_{q\big(x_0^{(k)}, \Ntk\big)} \left[ w_t\big(x_t^{(k)}, x_0^{(k)}\big)\, D_{\Phi, x_t^{(k)}}\Big(q_t\big(\cdot \mid x_0^{(k)}\big) \,\Big\|\, q_\theta^{(k)}(\cdot\;; \Ntk)\Big) \right]
\end{align*}

With $X:=q_t(\cdot\mid x_0^{(k)})=\alpha_t x_0^{(k)}+(1-\alpha_t)\pi_t$ and $w(X):=w_t(x_t^{(k)},x_0^{(k)})$. Because each term of the sum has its own $q_\theta^{(k)}$, minimizing implies minimizing each term. Lemma~\ref{lem:bregman-mean} gives that each term of the sum is minimized at:

\begin{equation*}
q^{\star}_\theta = \bar X_w = \frac{\mathbb{E}_{ q(x_0^{(k)}\mid N_{t,k})}\big[w_t(x_t^{(k)},x_0^{(k)})\,(\alpha_t x_0^{(k)}+(1-\alpha_t)\pi_t)\big]}{\mathbb{E}_{ q(x_0^{(k)}\mid N_{t,k})}\big[w_t(x_t^{(k)},x_0^{(k)})\big]} = \alpha_t g^\star + (1-\alpha_t)\pi_t,
\end{equation*}

where:

\begin{equation*}
    g^{\star}(x_0^{(k)}) = \frac{(x_0^{(k)})^T \cdot \mathbb{E}_{ q(x_0^{(k)}\mid N_{t,k})}\big[w_t(x_t^{(k)},x_0^{(k)})\,x_0^{(k)}\big]}{\mathbb{E}_{ q(x_0^{(k)}\mid N_{t,k})}\big[w_t(x_t^{(k)},x_0^{(k)})\big]} = \frac{w_t\big(x_t^{(k)},\,x_0^{(k)}\,\big)\; q\big(x_0^{(k)} \mid N_{t,k}\big)}{\displaystyle\sum_{v\in\mathcal V} w_t\big(x_t^{(k)}, v\big)\; q\big(v \mid N_{t,k}\big)}
\end{equation*}
Since $\alpha_t\in(0,1)$, $g\mapsto\alpha_t g+(1-\alpha_t)\pi_t$ is injective, $g^\star$ is the unique minimizer of $p_\theta$, giving \eqref{eq:locality-gidd-target}.
\end{proof}

Choosing $\pi_t = m$ we can recover Masked Diffusion; we therefore have the following corollary:

\begin{corollary*}[MDM recovers the untilted posterior]
Under the restriction of Proposition~\ref{prop:locality-gidd}, when $\pi_t\equiv m$ (the constant, absorbing-state mixing distribution), the locality optimum reduces to
\begin{equation*}
p_\theta^\star\big(x_0^{(k)} \mid N_{t,k}\big) = q\big(x_0^{(k)} \mid N_{t,k}\big).
\end{equation*}
\end{corollary*}
\begin{proof}
Since $\pi_t\equiv m$ is constant in $t$, $\pi_t'=0$, so the weighting function reduces to $w_t(x_t^{(k)},x_0^{(k)}) = -\frac{\alpha_t'}{\alpha_t}\,\frac{(x_t^{(k)})^\top m}{q_t(x_t^{(k)}\mid x_0^{(k)})}$. The proposition's target \eqref{eq:locality-gidd-target} is only evaluated at masked positions, $x_t^{(k)}=m$, where $(x_t^{(k)})^\top m = 1$ and, since clean tokens satisfy $x_0^{(k)}\neq m$ always, $q_t(m\mid x_0^{(k)}) = \alpha_t\cdot 0 + (1-\alpha_t)\cdot 1 = 1-\alpha_t$ irrespective of $x_0^{(k)}$. Hence $w_t(m,x_0^{(k)}) = -\alpha_t'/(\alpha_t(1-\alpha_t))$ is the same constant for every $x_0^{(k)}\in\mathcal V$, and cancels between numerator and denominator:
\begin{equation*}
p_\theta^\star\big(x_0^{(k)}\mid N_{t,k}\big) = \frac{q\big(x_0^{(k)}\mid N_{t,k}\big)}{\sum_{v\in\mathcal V} q\big(v\mid N_{t,k}\big)} = q\big(x_0^{(k)}\mid N_{t,k}\big).
\end{equation*}
\end{proof}

\subsection{Equivariance in the GIDD framework}

We will show that if our model $p_\theta$ is equivariant under a $G$ action, then its GIDD loss will be invariant. Remember that an invariant model is one where:

\begin{equation*}
p_\theta\big(x_0^{(k)} \mid g\cdot x_t\big) = p_\theta\big(x_0^{(g^{-1}\cdot k)} \mid  x_t\big), \qquad \forall\, g \in G,\ \forall\, k,\ \forall\, x_t,\ \forall\, x_0^{(k)}\in\mathcal V.
\end{equation*}

Then the following is true:

\begin{proposition*}[Equivariance of the weight and of the Bregman divergence]
Let $g\in G$ and $k\in\{1,\dots,d\}$ be arbitrary, $x_0, x_t \in \mathcal V^d$. Crucially, assume $p_\theta$ satisfies equivariance \eqref{eq:equivariance}. Then, the following are also invariant:
\begin{enumerate}
\item[(a)] $w_t\big((g\cdot x_t)^{(g\cdot k)},\, (g\cdot x_0)^{(g\cdot k)}\big) = w_t\big(x_t^{(k)}, x_0^{(k)}\big)$.
\item[(b)] $q_\theta^{(g\cdot k)}\big(\cdot\;;\, g\cdot x_t\big) = q_\theta^{(k)}(\cdot\;; x_t)$.
\item[(c)] $q_t\big(\cdot \mid (g\cdot x_0)^{(g\cdot k)}\big) = q_t\big(\cdot \mid x_0^{(k)}\big)$.
\item[(d)] $D_{\Phi,\, (g\cdot x_t)^{(g\cdot k)}}\Big(q_t\big(\cdot \mid (g\cdot x_0)^{(g\cdot k)}\big)\,\Big\|\, q_\theta^{(g\cdot k)}(\cdot\;; g\cdot x_t)\Big) = D_{\Phi,\, x_t^{(k)}}\Big(q_t(\cdot\mid x_0^{(k)})\,\Big\|\, q_\theta^{(k)}(\cdot\;; x_t)\Big)$.
\end{enumerate}
In particular, the summand of \eqref{eq:per-token-loss-bregman} at coordinate $g\cdot k$ evaluated at $(g\cdot x_0, g\cdot x_t)$ coincides with the summand at coordinate $k$ evaluated at $(x_0, x_t)$.
\end{proposition*}

\begin{proof}
By \eqref{eq:seq-action-cancel} applied to $x=x_t$ and $x=x_0$,
\begin{equation}
(g\cdot x_t)^{(g\cdot k)} = x_t^{(k)}, \qquad (g\cdot x_0)^{(g\cdot k)} = x_0^{(k)}.
\label{eq:cancel-both}
\end{equation}

(a) $w_t(u,v)$ is a fixed function of two tokens $u,v\in\mathcal V$ through $\alpha_t,\pi_t,\pi_t'$, quantities that do not depend on the coordinate. Substituting \eqref{eq:cancel-both} into the left-hand side gives $w_t(x_t^{(k)}, x_0^{(k)})$, the right-hand side. Note that (a) does not use the equivariance hypothesis \eqref{eq:equivariance}: it only relies on the schedule being coordinate-homogeneous, which is already assumed in the multidimensional reformulation.

(c) Similarly, $q_t(\cdot\mid x_0^{(k)}) = \alpha_t x_0^{(k)} + (1-\alpha_t) \pi_t$ is a fixed function of $x_0^{(k)}$ alone; substituting \eqref{eq:cancel-both} gives the result.

(b) We have
\begin{equation*}
q_\theta^{(g\cdot k)}(\cdot\;; g\cdot x_t) = \alpha_t\, p_\theta\big(x_0^{(g\cdot k)} = \cdot \,\big|\, g\cdot x_t\big) + (1-\alpha_t) \pi_t.
\end{equation*}
By the equivariance hypothesis, $p_\theta(x_0^{(g\cdot k)}=\cdot \mid g\cdot x_t) = p_\theta(x_0^{(k)}=\cdot\mid x_t)$ as a function over $\mathcal V$, hence
\begin{equation*}
q_\theta^{(g\cdot k)}(\cdot\;; g\cdot x_t) = \alpha_t\, p_\theta\big(x_0^{(k)}=\cdot\mid x_t\big) + (1-\alpha_t)\pi_t = q_\theta^{(k)}(\cdot\;; x_t).
\end{equation*}

(d) $D_{\Phi,z}(a\|b)$ is determined by the pair of distributions $(a,b)$ and the coordinate $z$ (Lemma~\ref{lem:kl-is-bregman}). Substituting $z=(g\cdot x_t)^{(g\cdot k)}=x_t^{(k)}$, $a = q_t(\cdot\mid(g\cdot x_0)^{(g\cdot k)}) = q_t(\cdot\mid x_0^{(k)})$ from (c), and $b = q_\theta^{(g\cdot k)}(\cdot\;; g\cdot x_t) = q_\theta^{(k)}(\cdot\;; x_t)$ from (b), the left-hand side reduces syntactically to the right-hand side, and we are done.
\end{proof}

Let us rewrite our loss in the following way:

\begin{equation*}
    \mathcal{L}_ t(\theta) = \mathbb{E}_{x_0, x_t}\big[\mathcal{L}_\theta(x_0,x_t)\big]=\mathbb{E}_{x_0, x_t} \Big[ \sum_{k=1}^d \ell^\theta_k(x_0, x_t)\Big]
\end{equation*}

where

\begin{equation*}
\ell^\theta_k(x_0, x_t) = w_t\big(x_t^{(k)}, x_0^{(k)}\big)\, D_{\Phi, x_t^{(k)}}\Big(q_t\big(\cdot \mid x_0^{(k)}\big) \,\Big\|\, q_\theta^{(k)}(\cdot\;; x_t)\Big).
 \end{equation*}

As a corollary from the latter proposition we have:

\begin{corollary*}[Invariance of the total loss]
Let $\mathcal{L}_\theta(x_0,x_t) := \sum_{k=1}^d \ell_k^\theta(x_0,x_t)$. If $p_\theta$ satisfies group equivariance, then
\begin{equation*}
\mathcal{L}_\theta(g\cdot x_0,\, g\cdot x_t) = \mathcal{L}_\theta(x_0,x_t) \qquad \forall\, g\in G,\ \forall\, x_0,x_t\in\mathcal V^d.
\end{equation*}
\end{corollary*}
\begin{proof}
Because $\ell^\theta_k$ is a product of invariant quantities ($w_t$ and $D_{\Phi, x_t^{(k)}}$) it is trivially invariant as well. Finally, as $\mathcal{L}_\theta$ is therefore a sum of invariant quantities, $\mathcal{L}_\theta(x_0, x_t) = \sum_{k=1}^d \ell^\theta_k(x_0,x_t)$, it is also invariant. 
\end{proof}

Therefore we can apply the invariance/equivariance results seen in the previous appendix section. And, if we pick $G$ to be the translation group $\mathcal{T}$, our GIDD local optimum will be agnostic to coordinates. In particular for the MDM case we have:

\begin{corollary*}[Translation-equivariant MDM optimum]
For a local-equivariant MDM and $G$ the translation group $\mathcal{T}$, the loss optimum is
\begin{equation}
p_\theta^{\star}\big(x_0^{(k)}{=}c \mid \Nt\big)
= \frac{\displaystyle\sum_{N_0 \in \Ndata} \ind{\mathrm{center}(N_0){=}c}\,\ind{N_0^{\mathrm{vis}}{=}\Nt^{\mathrm{vis}}}}
       {\displaystyle\sum_{N_0' \in \Ndata} \ind{N_0'^{\mathrm{vis}}{=}\Nt^{\mathrm{vis}}}},
\label{eq:mdm-local-equiv-opt-cor}
\end{equation}
where $\Nt^{\mathrm{vis}}$ denotes the restriction of $\Nt$ to its unmasked positions, undefined whenever no $N_0\in\Ndata$ agrees with $\Nt^{\mathrm{vis}}$ on every visible position.
\end{corollary*}
\begin{proof}
Substituting $\pi_t = m$ and using the dataset-amplification result from Appendix~\ref{app:theory-udm} we have our result. The forward corruption is the following:
\begin{equation*}
q(\Nt\mid N_0) = \alpha_t^{\,|\mathrm{vis}(\Nt)|}\,(1-\alpha_t)^{\,w^2-|\mathrm{vis}(\Nt)|}\;\ind{N_0^{\mathrm{vis}}{=}\Nt^{\mathrm{vis}}}.
\end{equation*}
The prefactor $\alpha_t^{|\mathrm{vis}(\Nt)|}(1-\alpha_t)^{w^2-|\mathrm{vis}(\Nt)|}$ depends only on $t$ and $\Nt$ and cancels when applying Bayes' theorem:
\begin{equation*}
q_{\mathrm{emp}}(x_0^{(k)}{=}c\mid\Nt) = \frac{\sum_{N_0\in\Ndata}\ind{\mathrm{center}(N_0)=c}\,q(\Nt\mid N_0)}{\sum_{N_0'\in\Ndata} q(\Nt\mid N_0')},
\end{equation*}
leaving exactly \eqref{eq:mdm-local-equiv-opt-cor}.
\end{proof}

For this paper we only present theoretical results for this framework, which offer great promise for future work extension in interpolating corruption frameworks.

\end{document}